\documentclass[preprint,12pt]{elsarticle}

\usepackage{amssymb}
\usepackage{url}
\usepackage{lineno,hyperref} 
\usepackage{xcolor}
\usepackage{ulem}
\usepackage{graphicx}
\usepackage{subfigure}
\usepackage{overpic} 

\usepackage{multirow}
\usepackage[ruled]{algorithm2e}
\usepackage{algpseudocode}  
\usepackage{amsmath}  

\journal{Nuclear Physics B}

\newcommand{\zll}[1]{\textcolor{black}{#1}}  
\newcommand{\lz}[1]{\textcolor{black}{#1}} 
\newcommand{\MajorMod}[1]{\textcolor{black}{#1}} 

\begin{document}
	
	\begin{frontmatter}
		
		
		
		\title{Feature matching for multi-epoch historical aerial images}
		
		
		\author{Lulin Zhang\textsuperscript{1}, Ewelina Rupnik\textsuperscript{1}, Marc Pierrot-Deseilligny\textsuperscript{1}}
		\address{\textsuperscript{1}LASTIG, Universit{\'e} Gustave Eiffel, ENSG, IGN, F-94160 Saint-Mand{\'e}, France - (Lulin.Zhang, Marc.Pierrot-Deseilligny)@ensg.eu,\\ Ewelina.Rupnik@ign.fr}
		
		%\affiliation{organization={},%Department and Organization
		
\begin{abstract}
	Historical imagery is characterized by high spatial resolution and stereoscopic acquisitions, providing a valuable resource for recovering 3D land-cover information. \MajorMod{Accurate geo-referencing of diachronic historical images by means of self-calibration} remains a bottleneck because of the difficulty to find sufficient amount of feature correspondences under evolving landscapes.	In this research, we present a fully automatic approach to detecting feature correspondences between historical images taken at different times (i.e., inter-epoch), without auxiliary data required. 
	Based on relative orientations computed within the same epoch (i.e., intra-epoch), we obtain DSMs (Digital Surface Model) and incorporate them in a rough-to-precise matching.
	% \lz{(By 3D geometry, we mean DSM (Digital Surface Model) for terrain geometry, and depth map for surfaces in image geometry.)} 
	The \lz{method consists} of: (1) an inter-epoch DSMs matching \lz{to roughly co-register the orientations and DSMs} (i.e, the \MajorMod{3D} Helmert transformation), followed by (2) a precise inter-epoch feature matching using the original \lz{RGB} images. The innate ambiguity of the latter is largely alleviated by narrowing down the search space using the co-registered \lz{data}.
	With the inter-epoch features, we refine the image orientations and quantitatively evaluate the results (1) with DoD ({\textit{Difference of DSMs}}), (2) with ground check points, and (3) by quantifying ground displacement due to an earthquake. We demonstrate that our method: (1) can automatically georeference diachronic historical images; (2) can effectively mitigate systematic errors induced by poorly estimated camera parameters; (3) is robust to drastic scene changes. Compared to the \textit{state-of-the-art}, our method improves the image georeferencing accuracy by a factor of 2. {The proposed methods are implemented in MicMac, a free, open-source photogrammetric software.}\\
	%with Digital Elevation Models (DEM) of differences (abbreviated as DoD) or ground displacement on 3 datasets, which demonstrated that our method effectively mitigated systematic errors in DoD, and leading to more accurate ground displacement.
\end{abstract}

		
		

	\end{frontmatter}
	
	

\section{Introduction}\label{MANUSCRIPT}
Historical imagery 
chronicles worldwide land-cover information, and as a result enables long-term environmental monitoring as well as 3D dynamic change detection. 
The images are of high spatial resolution, and are acquired in stereoscopic configuration. They have been acquired in many countries all over the world and can be traced back to the beginning of the 20$^{th}$ century~\cite{cowley2012historic}. \MajorMod{Millions} of historical images have been digitized and made accessible through web services~\cite{sebastien2019archiving,earthexplorer,remonterletemps}. They are often accompanied by metadata, in most cases including the camera focal length and the physical sensor size. Other metadata such as flight plans, camera calibration certificates or orientations are not commonly available. Besides,
inappropriate film/glass plate preservation and the scanning process enforce reestimating of the camera calibrations (i.e., the self-calibrating).\\
Self-calibration estimates simultaneously interior and exterior orientations. It is generally solved with a bundle block adjustment (BBA) routine, taking feature correspondences and Ground Control Points (GCPs) as input observations. Extracting feature correspondences within a single epoch (also referred to as \textit{intra-epoch}) can be efficiently done with local features such as SIFT~\cite{lowe2004distinctive}. Yet, due to drastic scene changes and heterogeneous acquisition conditions, it is challenging to automatically find feature correspondences across different epochs (also referred to as \textit{inter-epoch}).\\
In this work we propose a fully automatic approach to computing dense and robust inter-epoch feature correspondences. 
Our method consists of a rough co-registration by finding feature correspondences between DSMs derived within single epochs, and a precise feature matching on original RGB images. Our main contributions include:
\begin{itemize}
	\item \lz{By matching DSMs, we are able to obtain robust rough co-registration as the 3D landscape often stays globally stable over time.}
	\item \lz{\MajorMod{Under the guidance of co-registered orientations and DSMs, we reduce the difficulty in precise matching by:} (1) narrowing down the search space; (2) reducing the combinatorial complexity since only overlapping images are analyzed;}
	\item By proposing a tiling scheme (including \textit{one-to-many tiling} in rough co-registration and \textit{one-to-one tiling} in precise matching), we are opening up the possibility to scale-up the deep learning methods for feature matching. As we have shown in Section 4.3, using them out-of-the-box is inefficient. Large images demand high computing resources, while deep learning feature extraction methods are presumably trained on small images.
	\item \lz{By including priors about the geometry (in form of DSMs), we can filter candidate correspondences: each three correspondences projected to DSM serve to compute a 3D Helmert transformation between epochs, and most importantly provide a 2D constraint on all images' correspondences.}
\end{itemize}
In the following we briefly describe \MajorMod{the current states of local features, inter-epoch historical images alignment as well as robust matching}. In Section 3 we introduce our methodology, and in Section 4 the experiments as well as results are given. The source code is available from MicMac Github~\cite{micmacGithub}.\\
\section{Related Work}\label{MANUSCRIPT} 
\subsection{Local features}\label{sec:General Instructions}
Among hand-crafted features, SIFT~\cite{lowe2004distinctive} is undoubtedly a milestone. For decades, SIFT and its variants such as RootSIFT~\cite{arandjelovic2012three}, RootSIFT-PCA~\cite{bursuc2015kernel}, and DSP-SIFT~\cite{dong2015domain} have been widely used both in academic and industrial fields. \MajorMod{Other popular traditional features include ASIFT ~\cite{yu2011asift}, SURF ~\cite{bay2006surf}, Hessian Affine keypoint detector ~\cite{mikolajczyk2004scale} and KAZE ~\cite{alcantarilla2012kaze}. They are respectively powerful in affine invariance, computational efficiency, scale and affine invariance, retaining object boundaries.}\\
With the rise of deep learning, learned features have gained much attention due to their good performance on certain benchmarks. 
LIFT ~\cite{yi2016lift} is the first end-to-end network that implemented a full pipeline including detection, orientation estimation and feature description.
From then on, a large number of learned methods related to feature matching proliferated, including:
\begin{enumerate}
	\item Pure descriptor such as L2-Net~\cite{tian2017l2}, HardNet~\cite{mishchuk2017working} and Contextdesc ~\cite{luo2019contextdesc}.
	\item Detector along with descriptor. Most of the methods fall into this category, including  LF-Net~\cite{ono2018lf}, DELF~\cite{noh2017DELF}, SuperPoint~\cite{detone2018superpoint}, D2-Net~\cite{dusmanu2019d2},  ASLFeat~\cite{luo2020aslfeat}, R2D2~\cite{revaud2019r2d2}, D2D~\cite{wiles2020d2d}, etc.
	\item Matching methods based on off-the-shelf features, such as SuperGlue~\cite{sarlin2020superglue}. Originally combined with SuperPoint, SuperGlue was designed for real-time processing, and unlike the previous methods, it is capable of implicitly encoding spatial relation between neighboring features within an image and across images with the so-called attention mechanism.
\end{enumerate}
Early learned methods (LIFT ~\cite{yi2016lift}, L2-Net~\cite{tian2017l2}, HardNet~\cite{mishchuk2017working}, DELF ~\cite{noh2017DELF}, SuperPoint ~\cite{detone2018superpoint}, LF-Net ~\cite{ono2018lf}) use only intermediate metrics (e.g., repeatability, matching score, mean matching accuracy, etc.) to evaluate the matching performance. However, good performance on benchmarks does not necessarily imply a better matching quality ~\cite{schonberger2017comparative}.
Jin et al.~\cite{jin2020image} introduced a comprehensive benchmark for local features and robust estimation algorithms, focusing on the accuracy of the reconstructed camera pose as the primary metric. Using the new metric, SIFT ~\cite{lowe2004distinctive} and SuperGlue ~\cite{sarlin2020superglue} take the lead~\cite{imagematchingchallenge2020}.

\subsection{Inter-epoch historical images alignment}\label{sec:General Instructions}
When it comes to inter-epoch historical images, however, directly applying SIFT or SuperGlue often results in inferior results due to large radiometric differences.
In Figure~\ref{comparison} we showed an example where SIFT and SuperGlue failed on an inter-epoch image pair with drastic scene changes. It is understandable as (1) SIFT is not sufficiently invariant over time, while (2) SuperGlue is not invariant to rotations and it underperforms on larger images because it was presumably trained on small images.\\
Therefore, many previous researches bypassed the task of extracting inter-epoch correspondences by processing different epochs separately followed by an inter-epoch co-registration relying on Ground Control Points(GCPs).
\MajorMod{Between 10 and 169 GCPs are required in ~\cite{pinto2019archived},~\cite{bozek2019analysis},~\cite{persia2020archival},~\cite{micheletti2015application},~\cite{molg2017structure}.}
GCPs are usually measured with the help of photointerpretation on recent orthophotos, however, it is still monotonous and time-consuming. Furthermore, it is difficult to find salient points that are stable over time.\\
Certain attempts were made to extract inter-epoch correspondences. Giordano et al.~\cite{giordano2018toward} extract feature correspondences between historical and recent images relying on HoG descriptors~\cite{dalal2005histograms}. The authors require flight plans as input, which are not commonly available as mentioned in Section 1. Feurer et al.~\cite{feurer2018joining}, Filhol et al.~\cite{filhol2019time}, Cook et al.~\cite{cook2019simple}, Parente et al.~\cite{parente2021automated} and Blanch et al.~\cite{blanch2021multi} assume that a sufficient number of keypoints remain invariant across time and employ SIFT to extract inter-epoch feature correspondences. It remains questionable whether the method is capable of handling drastic scene changes.
Zhang et al.~\cite{zhang2020guided} extract inter-epoch correspondences from SIFT-detected keypoints based on the hypothesis that points follow 2D and 3D spatial similarity model. This method works in simple cases with few scene changes.
%\sout{Maiwald et al.~\cite{maiwald2021automatic} use D2-Net to match historical terrestrial urban images for the purpose of knowledge transfer in cultural heritage. However it is shown in ~\cite{zhang2020guided} that D2-Net provides keypoints of low accuracy, therefore is not suitable for metrological applications.}
{Additionally, a stream of research works focuses on historical terrestrial images (~\cite{maiwald2021automatic}, ~\cite{beltrami20193d}, ~\cite{bevilacqua2019reconstruction}, ~\cite{maiwald2019generation}) and historical video recordings (~\cite{maiwald2019generation}). However, their algorithms are not suitable to the aerial case.}\\
This work is an extention of~\cite{zhang2020guided}. Unlike in~\cite{zhang2020guided}, we introduce a rough co-registration between different epochs based on matching DSMs with SuperGlue, and use it to guide a precise matching. Our rough co-registration is robust under extreme scene changes because (1) SuperGlue utilizes context to enhance feature descriptors and (2) DSMs are generally stable over time. With the guidance of roughly co-registered orientations and DSMs, both SIFT and SuperGlue achieved good performance, as shown in our experiments.
\subsection{Robust matching}\label{sec:General Instructions}
The goal of robust matching is to tell apart inliers \zll{(i.e., true correspondences)} from outliers \zll{(i.e., false correspondences)}, and eliminate the latter from further processing.
Typically, an iterative sampling strategy based on RANSAC (Random Sample Consensus) ~\cite{fischler1981random} relying on some geometric model, such as homography ~\cite{sonka2014image} or essential matrix ~\cite{sonka2014image} is carried out to remove outliers. 
Substitutes of RANSAC include hand-crafted methods such as LMedS~\cite{leroy1987robust} (a meaningful groundwork before RANSAC), MLESAC ~\cite{torr2000mlesac} (maximizing inlier likelihood), PROSAC~\cite{chum2005matching} (choosing samples progressively for acceleration), DEGENSAC ~\cite{chum2005two} (increasing robustness when a dominant plane is present), GC-RANSAC ~\cite{barath2018graph} (adopting graph-cut algorithm), MAGSAC ~\cite{barath2019magsac} (defining threshold automatically with marginalization), and learned methods such as DSAC~\cite{brachmann2017dsac} (simulating RANSAC in a differentiable way) and CNe~\cite{moo2018learning} (end-to-end deep network).\\
In this research, we use RANSAC to estimate the \MajorMod{3D} Helmert transformation between surfaces (i.e., DSMs) calculated in different epochs. Compared to the classical essential/fundamental matrix filtering, with less data (3 versus 5 points) we impose stricter rules on the sets of points. Lastly, we eliminate the remaining false correspondences by looking at their cross-correlation.
\section{Methodology}
\begin{figure*}[htbp]
	\begin{center}
		\subfigure[Full processing workflow]{
			\begin{minipage}[t]{1\linewidth}
				\centering
				\includegraphics[width=13cm]{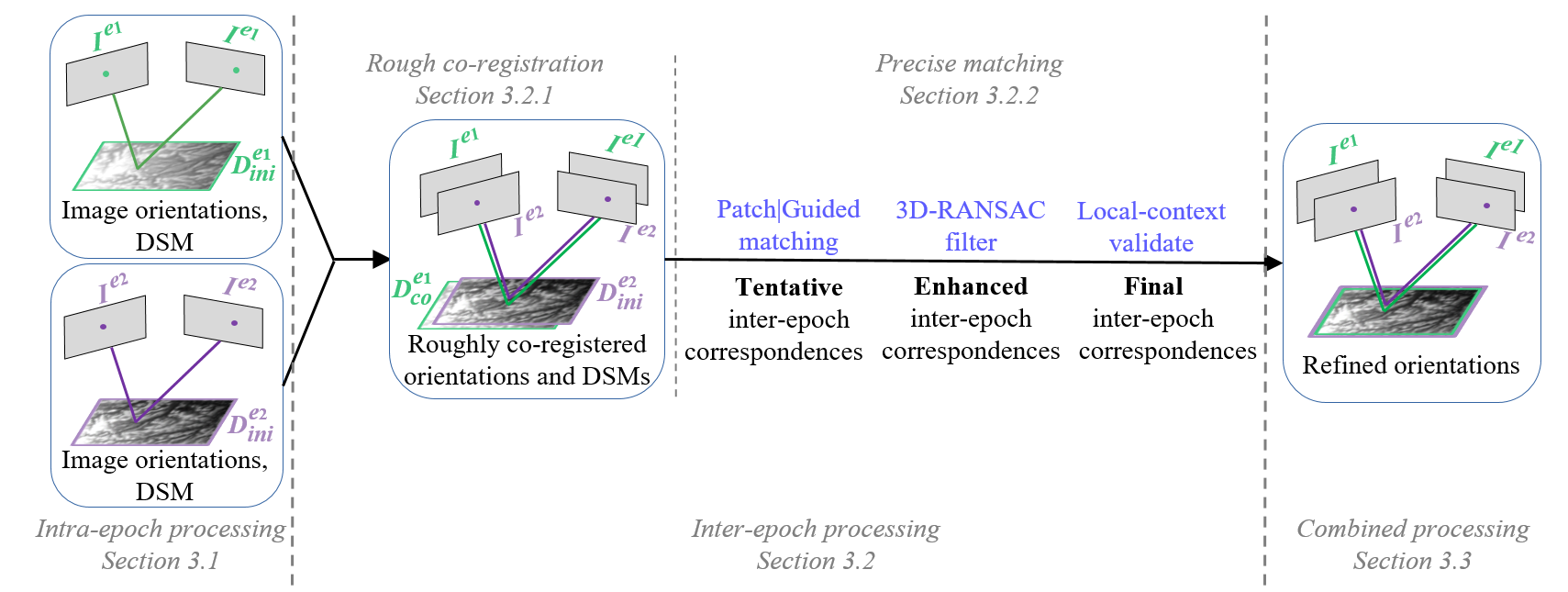}
			\end{minipage}%
		}
		\subfigure[\MajorMod{Workflow of the rough co-registration.}]{
	\begin{minipage}[t]{1\linewidth}
		\centering
		\includegraphics[width=13cm]{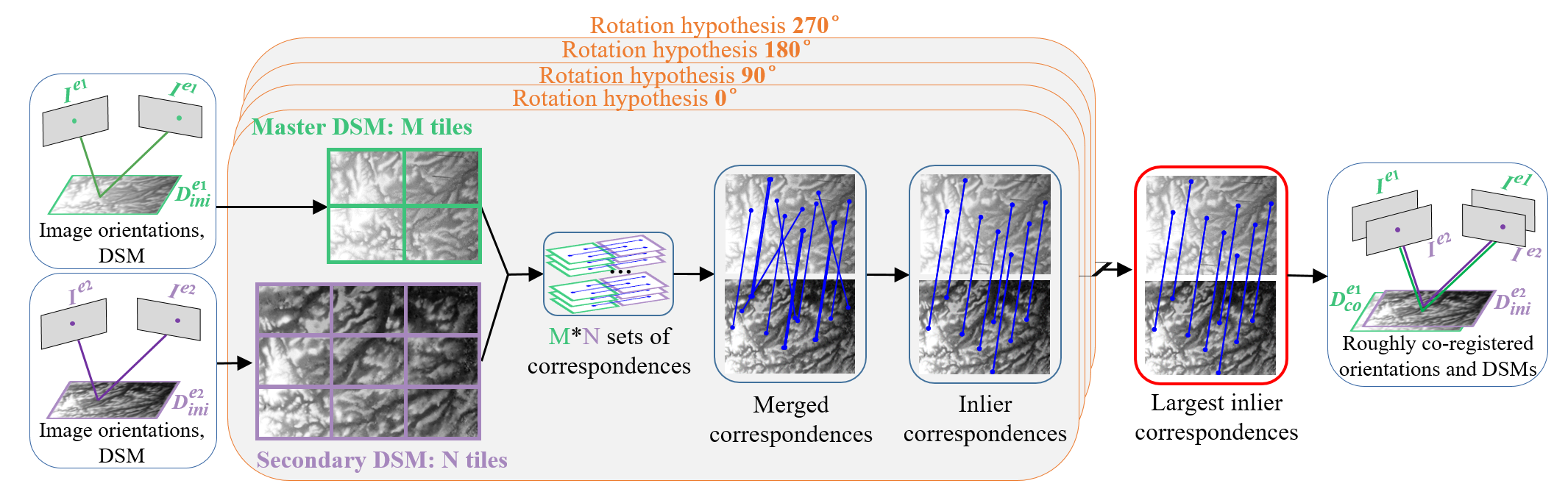}		
	\end{minipage}%
	}
		\subfigure[Four rotation hypotheses]{
	\begin{minipage}[t]{1\linewidth}
		\centering
		\includegraphics[width=13cm]{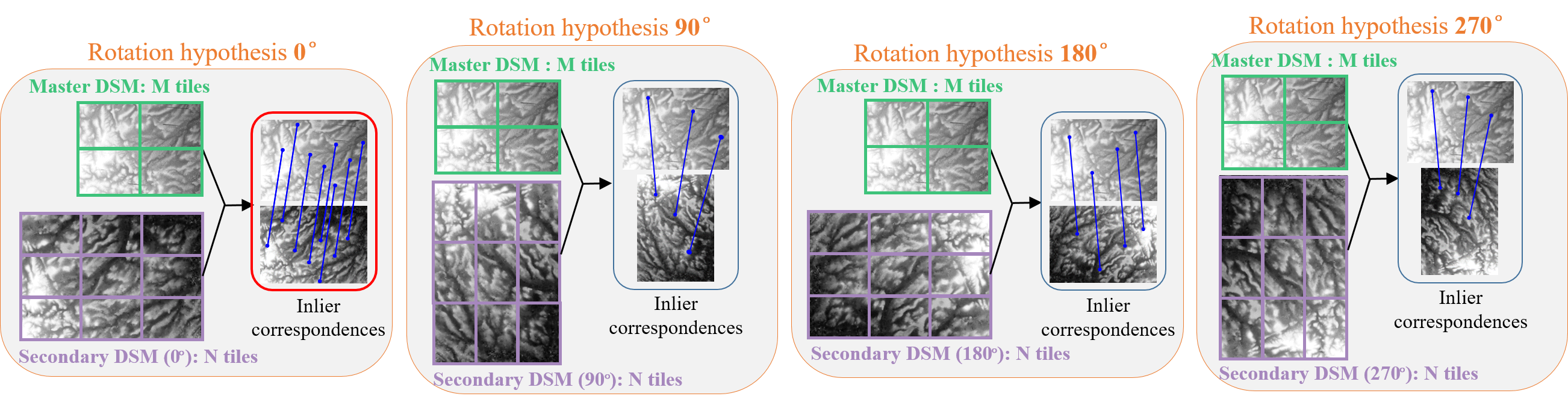}
	\end{minipage}%
	}
		\caption{(a) Our full processing workflow. Intra-epoch feature correspondences, relative orientations and DSM are obtained within each epoch individually. Once multi-epochs are roughly co-registered \MajorMod{based on feature correspondences} computed on DSMs, a precise matching is carried out by performing patch matching or guided matching, followed by a 3D-RANSAC filter and a local-context validation, giving rise to the final inter-epoch feature correspondences. (b) Workflow of the rough co-registration. SuperGlue is applied on tiles of DSMs, followed by a RANSAC procedure to remove outliers. As SuperGlue is not invariant to rotations lager than 45$^{\circ}$, \MajorMod{we test four rotation hypotheses} and keep the best one. \MajorMod{(c) Four rotation hypotheses. We rotate the secondary DSM by 90$^{\circ}$ four times to match with master DSM and keep the best one with the largest number of RANSAC inliers.}}
		\label{Flow-process diagram}
	\end{center}
\end{figure*}
Figure~\ref{Flow-process diagram}(a) exhibits the workflow of our pipeline.
It includes 3 main parts: intra-epoch processing, inter-epoch processing, and combined processing. \lz{The images' focal lengths and physical sensor sizes are supposed to be known, as we already mentioned in Section 1}. The images are resampled to the geometry of the fiducial marks prior to processing. 
For the sake of simplicity, we only exhibit the processing flow of two epochs, however, it can be easily extended to more epochs.\\
\MajorMod{We use the term \textit{Patch} and \textit{Guided} to refer to two alternatives of our rough-to-precise matching pipeline (patch matching or guided matching). The patch matching uses learned features, and the guided matching uses hand-crafted features. We explore the two approaches because of their diverse characteristics: the \textit{Patch} recovers more correspondences (cf. Figure~\ref{comparison} (b) and (c)), however, due to the resampling stage, they are computationally more intensive compared to \textit{Guided}. Additionally, our collateral goal is to compare their performances.}\\
We adopt the following naming conventions -- (1) images acquired in \textit{epoch$_1$} are denoted as $I^{e_1}$, and images in \textit{epoch$_2$} as $I^{e_2}$; (2) orientations of \textit{epoch$_1$} and \textit{epoch$_2$} are denoted as $O^{e_1}$ and $O^{e_2}$; (3) DSMs of \textit{epoch$_1$} and \textit{epoch$_2$} are denoted as $D^{e_1}$ and $D^{e_2}$.
\subsection{Intra-epoch Processing}
\lz{Intra-epoch processing furnishes complementary 3D information to help the latter inter-epoch processing. It is a standard photogrammetry or SfM pipeline and can be accomplished with lots of solutions (e.g. MicMac~\cite{deseilligny2011apero}, COLMAP~\cite{schonberger2016structure}, {OpenMVG~\cite{openMVG}, Theia~\cite{theia}, etc.}). The solution used in our experiment is MicMac.} It is performed within each \textit{epoch$_i$} individually as follows:
\begin{enumerate}
	\item Extract intra-epoch correspondences between images $I^{e_i}$ with SIFT ~\cite{lowe2004distinctive};
	\item Compute interior and relative orientations ({$O_{ini}^{e_i}$}) with the sequential SfM;
	\item Based on image orientations $O_{ini}^{e_i}$, perform semi-global dense matching~\cite{mpd:06:sgm} {between images $I^{e_i}$} to get {DSM ($D_{ini}^{e_i}$) in their arbitrary coordinate frames.}
\end{enumerate}
\subsection{Inter-epoch Processing}
Inter-epoch processing follows a rough-to-precise matching strategy including:\\ 
\begin{enumerate}
	\item Rough co-registration: match DSMs based on \textit{one-to-many tiling} scheme (described in Section 3.2.1) to roughly co-register the image orientations and DSMs into a common reference frame.\\
	\item Precise matching: match original RGB images under the guidance of the roughly co-registered data.\\
\end{enumerate}
\MajorMod{We choose matching DSMs for rough co-registration and RGB images for precise matching because: 
 (1) DSMs computed from historical images turn to be more constant over time compared to RGB images; however, (2) due to low radiometric quality of the images, the DSMs are noisy (cf. Figure~\ref{precisematchingdepth}(d)).
As a consequence, DSMs provide correspondences perfectly tailored for rough co-registration - less accurate but repetitive over time. Then, by narrowing down the search space based on co-registered orientations, matching RGB images will give us correspondences that are both accurate and robust.
}
\subsubsection{Rough co-registration}
{The workflow of rough co-registration is displayed in Figure~\ref{Flow-process diagram}(b).}
The results (co-registered orientations and DSMs) will be used to (1) guide the precise matching, and (2) {provide} initial orientations for bundle adjustment in the combined processing. Therefore, the goal is to get a moderate number of reliable inter-epoch feature correspondences at low cost.
We choose SuperGlue~\footnote{We use the pre-trained SuperGlue model provided by the authors. Keypoint locations as well as confidence scores are predicted with SuperPoint~\cite{detone2018superpoint}.} to do the rough matching as it is more invariant over time than SIFT.
Simply applying SuperGlue on multi-epoch images fails because it is not rotation invariant and it underperforms on large images. \MajorMod{We improve the matching robustness by adopting the following modifications:}
\paragraph{(1) Match DSMs instead of original images}
{Assuming the numbers of the original images in \textit{epoch$_1$} and \textit{epoch$_2$} are P and Q respectively}, instead of applying brute force rough matching P$\times$Q times on original images, we only need to do it once on a pair of DSM images.
To obtain the DSM images, we convert the DSMs (i.e., 2.5D rasters) from floating-point to [0-255] range grayscale images. Since the DSMs contain outliers, pixels with elevations larger than double the standard deviation of all elevations are ignored in the conversion process.
Matching DSM images has the following merits:
\begin{enumerate}
	\item Redundancy caused by the forward and side overlapping areas is removed;
	\item It {implicitly enables} a follow-up search for globally consistent inliers;
	\item It decreases the combinatorial complexity caused by rotation ambiguity of the product of P and Q images;
	\item {Even under important scene changes, DSMs generally provide stable information over time.}
\end{enumerate}
\paragraph{(2) One-to-many tiling scheme}
As SuperGlue provides unsatisfactory result on large images, we propose a \textit{one-to-many tiling scheme} to make up for the deficiency. It is performed as follows (cf. Figure~\ref{Flow-process diagram}(b)):
\begin{enumerate}
	\item Crop both the master and secondary DSM images into M and N tiles of certain size, respectively;
	\item Apply SuperGlue on M$\times$N tile pairs individually;
	\item Merge the feature correspondences and perform RANSAC based on 2D similarity transformation to find globally consistent inliers.
\end{enumerate}
\paragraph{(3) Testing 4 rotation hypotheses}
As SuperGlue is not invariant to rotations larger than 45$^{\circ}$, the \textit{one-to-many tiling scheme} will be performed repeatedly {over} 4 rotation hypotheses as follows (cf. Figure~\ref{Flow-process diagram}(c)): 
\begin{enumerate}
	\item Rotate the secondary DSM image by 90$^{\circ}$ four times;
	\item Match each rotated DSM image with the master DSM image with the \textit{one-to-many tiling scheme};
	\item Keep the hypothesis with the largest number of RANSAC inliers.
\end{enumerate}
With RANSAC inliers, we build a \MajorMod{3D} Helmert transformation to transform the orientations $O_{ini}^{e_1}$ and DSM $D_{ini}^{e_1}$ to the frame of ${epoch_2}$, leading to co-registered orientations $O_{co}^{e_1}$ and DSM $D_{co}^{e_1}$.\\
The feature matching method based on DSMs will fail on perfectly flat terrain. In this rare scenario, feature correspondences can be extracted on orthophotos {(see demonstration in Section 4.3)}, and their elevation coordinates can be retrieved from the DSMs. However, because the scene radiometry change {is more pronounced} than the 3D landscape change, fewer matches are found.\\
\subsubsection{Precise matching}
\begin{figure*}[h!]
	\begin{center}
		\subfigure[patch matching]{
			\begin{minipage}[t]{0.31\linewidth}
				\centering
				\includegraphics[width=3.7cm]{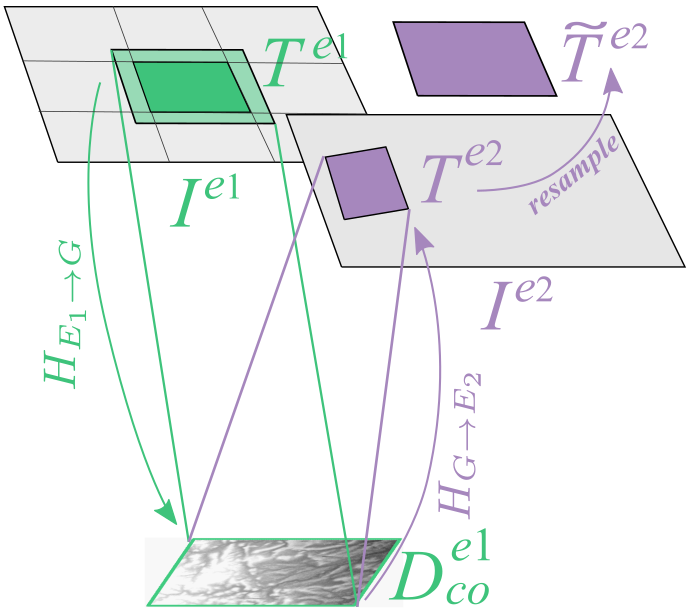}
				
			\end{minipage}%
		}
		\subfigure[Buffer zone]{
			\begin{minipage}[t]{0.31\linewidth}
				\centering
				\includegraphics[width=3.5cm]{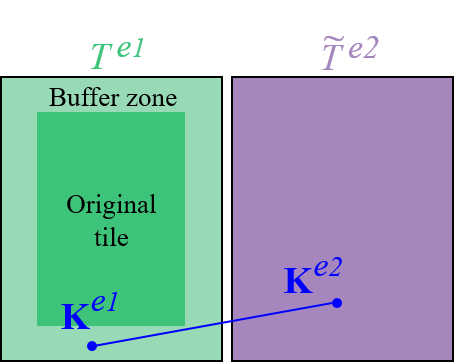}
				
			\end{minipage}%
		}
		\subfigure[guided matching]{
			\begin{minipage}[t]{0.31\linewidth}
				\centering
				\includegraphics[width=3.3cm]{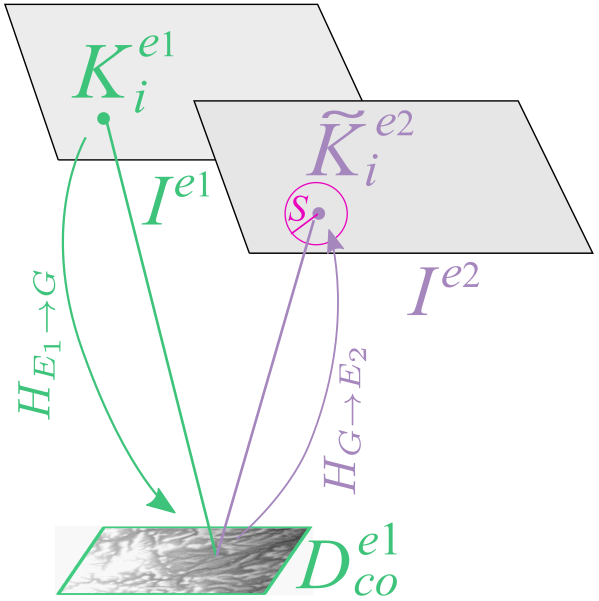}
				
			\end{minipage}%
		}
		\caption{(a) and (c) illustrate toy-examples of the patch matching and guided matching, respectively, (b) displays the feature correspondences where $\mathbf{K}^{e_1}$ exceeds the original tile size (dark green area) and therefore will be abandoned.}
		\label{patchmatch}
	\end{center}
\end{figure*}

\begin{figure*}[htbp]
	\begin{center}
		\includegraphics[width=0.8\columnwidth]{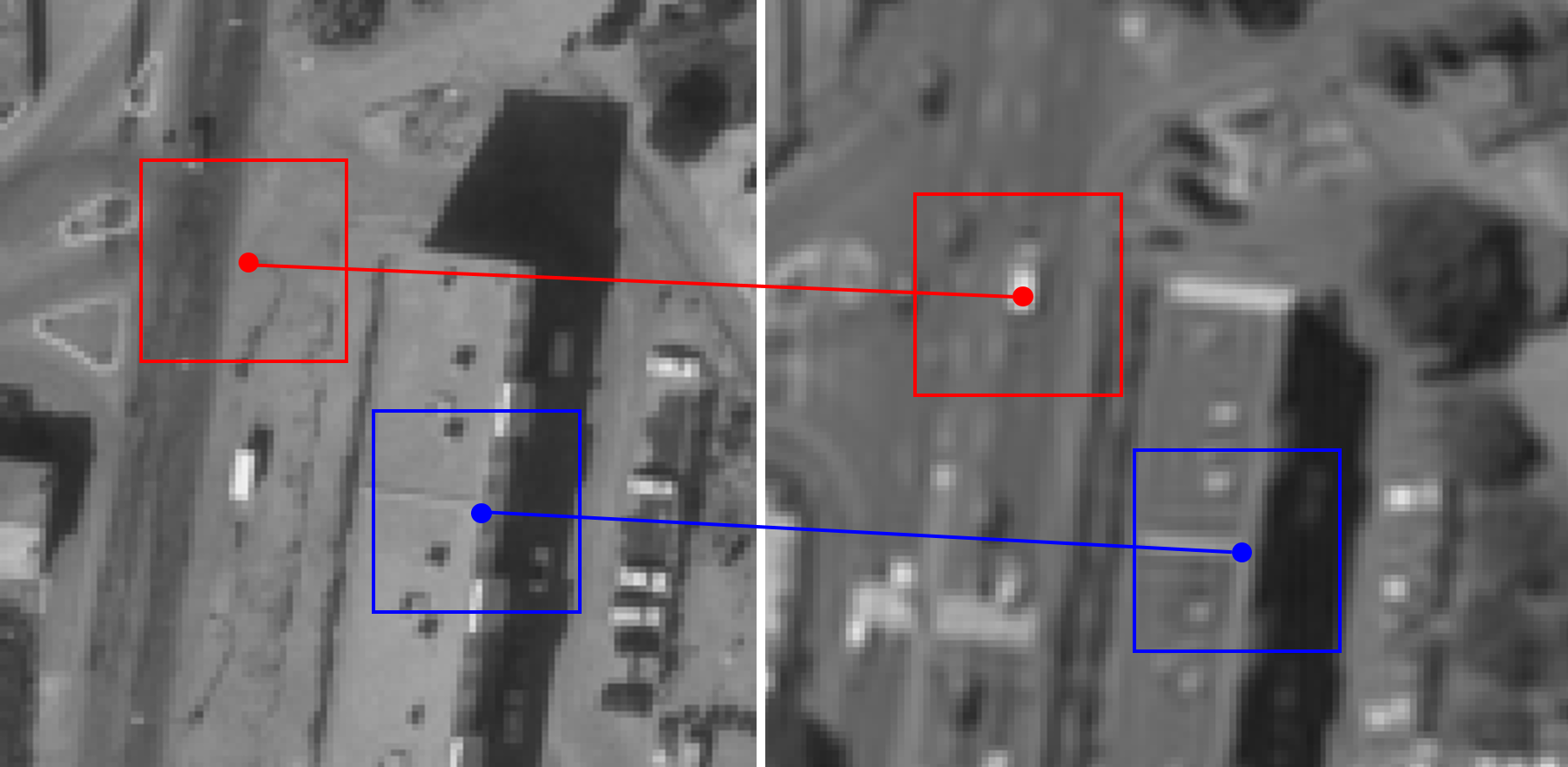}
		\caption{Demonstration of the validation with cross-correlation. Considering poor quality of historical images, the window size (blue and red rectangles) was set to 32$\times$32 pixels. False match (red) is eliminated by cross correlation, while true match (blue) is kept.}
		\label{crossc}
	\end{center}
\end{figure*}

To compute precise {inter-epoch} feature correspondences, we perform matching on original RGB images {under the guidance of co-registered orientations and DSMs}. {It consists of extracting tentative inter-epoch feature correspondences, followed by a 3D-RANSAC filter and a local-context validation stage to remove outliers.}
\MajorMod{
\paragraph{(1) Get tentative inter-epoch correspondences}\label{patch matching}
We offer two alternatives to get tentative correspondences: patch matching or guided matching. {The former has better overall} performance while the latter is more efficient in terms of the use of memory and CPU resources.\\
\textit{(1-1) Patch matching for learned features}
}
\MajorMod{{It} is based on \textit{one-to-one tiling scheme}} {(not to confuse with the \textit{one-to-many tiling scheme} presented in the \textit{rough matching}), as shown in Figure~\ref{patchmatch}(a) and detailed below}:
\begin{enumerate}
	\item Crop the master RGB image $I^{e_1}$ into M tiles ($T^{e_1}$) of certain size, with a buffer zone overlapped with each other;
	\item As the orientations and DSMs are roughly co-registered, we project each master tile $T^{e_1}$ onto the DSM $D_{co}^{e_1}$ and backproject to secondary RGB image $I^{e_2}$ to find the corresponding tile $T^{e_2}$;
	\item Resample $T^{e_2}$ to $\widetilde{T}^{e_2}$, so that the tile pair $P({T^{e_1},\widetilde{T}^{e_2}})$ is free from differences of rotation, scale and extent;
	\item Apply SuperGlue on tile pair $P({T^{e_1},\widetilde{T}^{e_2}})$ to find feature correspondences $M({\mathbf{K}^{e_1},\mathbf{K}^{e_2}})$ ($\mathbf{K}^{e_i}$ represents keypoints in $I^{e_i}$);
	\item Merge the feature correspondences.
\end{enumerate}
Our patch matching experiments are performed based on SuperGlue, however, other learned methods can be adopted readily. Because the orientations and DSMs are only roughly co-registered, we have to take into account the margin of error when projecting tiles to overlapping images. \MajorMod{This is why we add a buffer zone around the tile $T^{e_1}$,} as shown in Figure~\ref{patchmatch}(b). To avoid duplicate matches, we remove \textit{a posteriori} the matches exceeding the original tile.\\
\textit{(1-2) Guided matching for hand-crafted features.} The patch matching substitute orientated towards hand-crafted features is the guided matching, as shown in Figure~\ref{patchmatch}(c). It leverages \MajorMod{the positions of} predicted keypoints, {the known scale ratio and rotation differences to narrow down the list of the matching candidates}. In our experiments, we use the SIFT points, but the method is suitable to any hand-crafted extractor.
The strategy {is as follows}:\\
\begin{enumerate}
	\item {Compute the scale ratio $R_{scl}$ and the rotation $D_{rot}$ between two images by sequentially projecting the $I^{e_1}$ image corners to the co-registered DSM $D_{co}^{e_1}$ and to image $I^{e_2}$;} %\textcolor{red}{Project the image corners of $I^{e_1}$ to the co-registered DSM $D_{co}^{e_1}$, and back-project them to $I^{e_2}$ to estimate the scale ratio $R_{scl}$ and angle difference $D_{ang}$ between images $I^{e_1}$ and $I^{e_2}$.}
	\item Extract keypoints $\mathbf{K}^{e_1}$ in image $I^{e_1}$ and $\mathbf{K}^{e_2}$ in image $I^{e_2}$;
	\item Intersect the keypoints $\mathbf{K}^{e_1}$ with the co-registered DSM $D_{co}^{e_1}$;
	\item Back-project them to image $I^{e_2}$, giving rise to predicted keypoints $\widetilde{\mathbf{K}}^{e_2}$;
	\item Search for a subset of points in $\mathbf{K}^{e_2}$ located within a radius $S$ (100 pixels in our experiments) centered at the predicted positions $\widetilde{\mathbf{K}}^{e_2}$;%\textcolor{green}{(I don't use a distance threshold.)}
	\item {Remove candidate matches whose scales and rotations computed by SIFT are incoherent with $R_{scl}$ and $D_{rot}$ computed from image orientations and the co-registered DSM (i.e., step 1);}
	\item {Find the best match with mutual nearest neighbor and apply the first to second nearest neighbor ratio test~\cite{lowe2004distinctive}.}
\end{enumerate}
\paragraph{(2) Get enhanced inter-epoch correspondences}
To compute enhanced correspondences, we apply a 3D-RANSAC filter on the previously obtained tentative matches. {More precisely, we do the following}: (1) for each feature correspondence $M({\mathbf{K}^{e_1},\mathbf{K}^{e_2}})$, the keypoints $\mathbf{K}^{e_1}$ and $\mathbf{K}^{e_2}$ are projected onto DSM $D_{co}^{e_1}$ and $D_{ini}^{e_2}$ individually to get 3D points $M({\mathbf{G}^{e_1},\mathbf{G}^{e_2}})$; and (2) {the correspondences} $M({\mathbf{G}^{e_1},\mathbf{G}^{e_2}})$ are iteratively sampled to compute the 3D spatial similarity RANSAC model:
\begin{equation*}
{G_i^{e_2}} = \lambda \cdot \mathbf{R} \cdot {G_i^{e_1}} + \mathbf{T} , \quad i \in [1,3] \label{eq:3Dsim}
\end{equation*}
where $\lambda$ is the scale factor, $\mathbf{T}$ is the translation vector and $\mathbf{R}$ is the rotation matrix.
We set the number of RANSAC iterations to 1000, and consider feature correspondences within $T_r$ of its predicted position as inliers. In our experiment, {$T_r$ was set to 10$\times$$GSD$ where $GSD$ is the mean ground sampling distance in the coordinate frame of epoch ${e_2}$. This distance is computed as the ground distance between two adjacent image pixels.}
%0.01 unit {you should motivate this number, why 0.01?} of the relative and arbitrary coordinate system.
\paragraph{(3) Get final inter-epoch correspondences}
In the preceding step we got rid of a substantial number of outliers, however, we believe that not all outliers could be identified. We apply cross-correlation for final validation. Feature correspondences with their correlation scores below a predefined threshold (0.6 in our experiments) are discarded. The correlation window size was set to be large enough to take into account the context around a point (32$\times$32 pixels in our experiment). Figure~\ref{crossc} shows an example of a false match (red) eliminated by cross correlation, while the true match (blue) is kept.
\subsection{Combined Processing}
Based on the intra-epoch and inter-epoch feature correspondences, a free network BBA is performed to refine all the image orientations and camera calibrations. If the results need to be analyzed in a metric scale, a spatial similarity transformation will be performed to move the refined acquisitions in an arbitrary reference frame to a metric one. If the precise orientations for one of the epochs were known \MajorMod{(i.e., deemed as ground truth)}, their parameters will be fixed during the BBA and the subsequent spatial similarity transformation will be skipped.
We adopted the Fraser model ~\cite{fraser1997digital} to calibrate the cameras and allowed image-dependent affine parameters, the remaining parameters were
shared among all images.
\section{Experiment}
\subsection{Implementation details}
{To reduce the image noise} all historical images entering the pipeline are downsampled by a factor of 3. To calculate the DSMs, we further downsample the images by a factor of 4, {which amounts to a total downsampling factor of 12 with respect to the input images. For example, the images in Fr{\'e}jus 1970 are downsampled from [8766, 8763] to [730, 730]}. {{Note that} the DSMs serve 3 purposes: (1) rough co-registration, (2) narrowing down the search space in precise matching, (3) providing 3D coordinates for 3D-RANSAC filter. A low resolution surface is good enough for these tasks, and keeps the computational cost low.} \\
To balance the number of the intra- and inter-epoch feature correspondences, we perform intra-epoch feature correspondences reduction available in MicMac~\cite{marc2016micmac}, followed by setting the relative observation weight in the bundle adjustment, if necessary. The feature correspondences reduction algorithm maximizes good spatial distribution, points' multiplicity and low reprojection error, it also helps to speed up the bundle adjustment.\\
Inter-epoch feature correspondences are extracted {for every possible combination of 2 epochs and finally merged}.
Algorithm~\ref{alg:Framwork} demonstrates the pseudo code of our pipeline.\\
\begin{algorithm}[htbp] % \label{peseudocode1}
	\caption{ Our pipeline}  
	\label{alg:Framwork}  
	\LinesNumbered 
	%\begin{algorithmic}[1]  
	\KwIn{Images; focal lengths; physical sensor sizes;}
	\KwOut{Interior and relative orientations;} 
	\For{each epoch \textit{i}}{  
		Downsample historical images by a factor of 3;\\ 
		Extract intra-epoch feature correspondences with SIFT, followed by a reduction of feature correspondences, giving rise to $T_{intra}$;\\ 
		Recover interior and relative orientations \MajorMod{$O_{ini}^{e_i}$};\\
		Calculate DSMs \MajorMod{$D_{ini}^{e_i}$} based on images further downsampled by a factor of 4;\\
	}
	Set the most recent epoch as reference epoch $E_r$, the other epochs as free epoch $E_f$;\\
	\For{every epoch $E_f$}{
		Convert DSM \MajorMod{{$D_{ini}^{e_f}$}} and \MajorMod{$D_{ini}^{e_r}$} to grayscale images, leading to \MajorMod{$I_{DSM}^{e_f}$} and \MajorMod{$I_{DSM}^{e_r}$};\\
		Match \MajorMod{$I_{DSM}^{e_f}$} and \MajorMod{$I_{DSM}^{e_r}$} with \MajorMod{\textit{one-to-many tiling scheme}};\\
		Transform orientations \MajorMod{$O_{ini}^{e_f}$} and DSM \MajorMod{$D_{ini}^{e_f}$} to the frame of epoch $E_r$, leading to orientations \MajorMod{$O_{co}^{e_f}$} and DSM \MajorMod{$D_{co}^{e_f}$};\\
	}
	\For {each inter-epoch image pair}{
		Get tentative feature correspondences with patch or guided matching, based on \MajorMod{orientations $O_{co}$ and} DSM $D_{co}$;\\
		Get enhanced feature correspondences with 3D-RANSAC filter, based on \MajorMod{orientations $O_{co}$ and} DSM $D_{co}$;\\
		Get final feature correspondences (i.e., $T_{inter}$) with cross correlation;\\
	}
	Refine orientations $O_{co}$ with $T_{intra}$ and $T_{inter}$;\\
\end{algorithm}  
\subsection{Datasets}
\begin{figure*}[htbp]
	\begin{center}
		\subfigure[Pezenas]{
			\begin{minipage}[t]{0.35\linewidth}
				\centering
				\includegraphics[width=4.4cm]{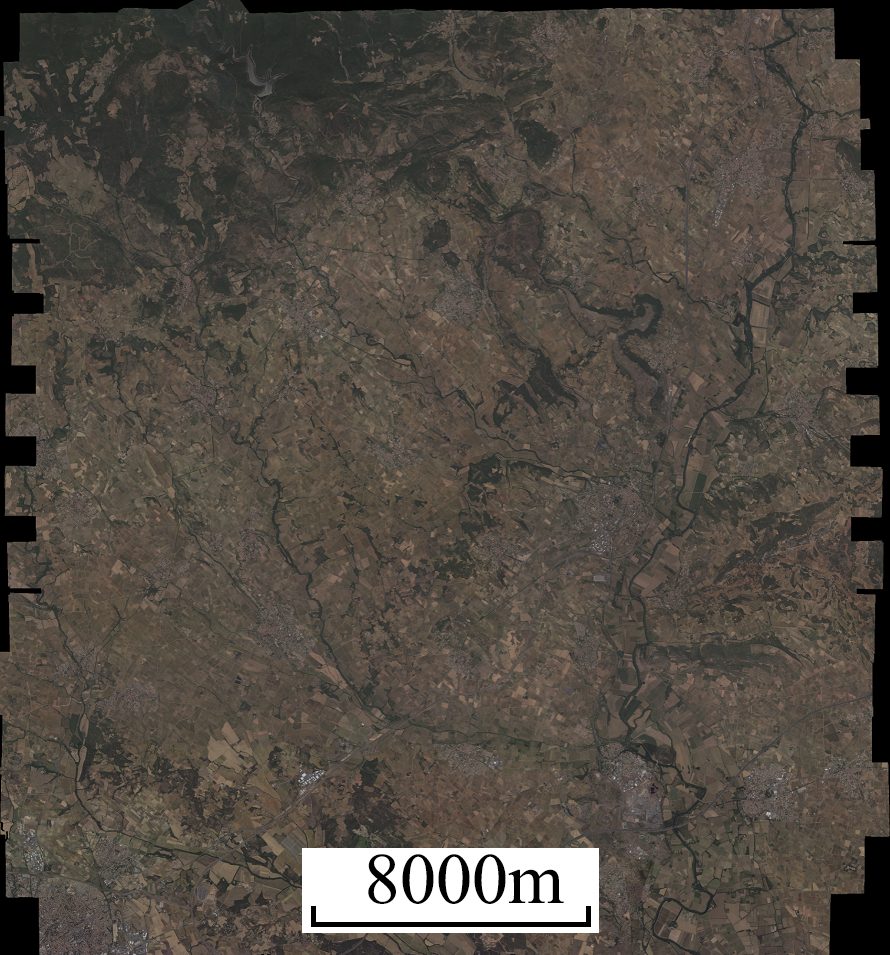}
				%\caption{DoD$_{Pezenas\_1954}^{Co-Reg}$}
			\end{minipage}%
		}
		\subfigure[Fr{\'e}jus]{
			\begin{minipage}[t]{0.62\linewidth}
				\centering
				\includegraphics[width=8cm]{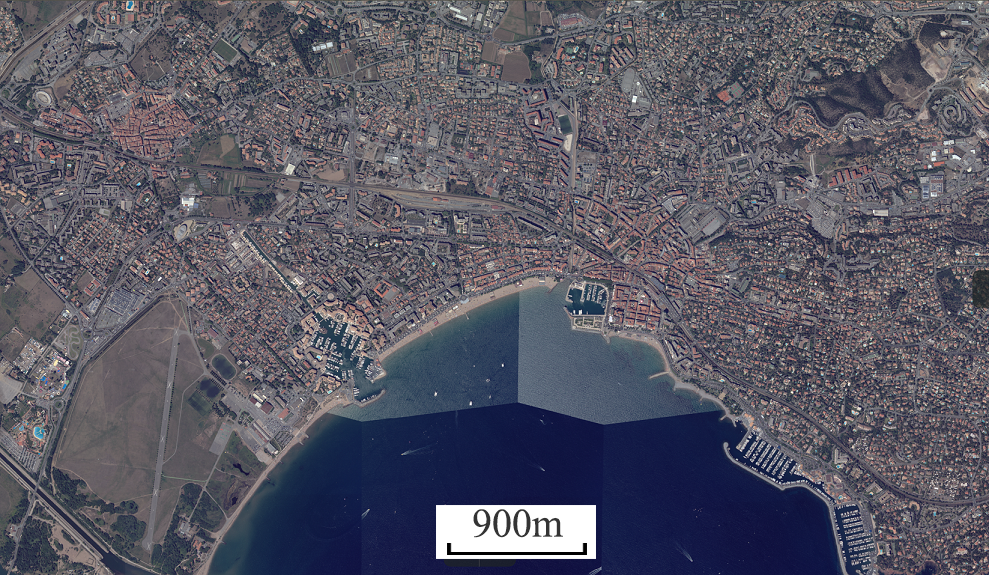}
				%\caption{DoD$_{Frejus\_1954}^{Co-Reg}$}
			\end{minipage}%
		}
		\subfigure[Kobe]{
			\begin{minipage}[t]{1\linewidth}
				\centering
				\includegraphics[width=13cm]{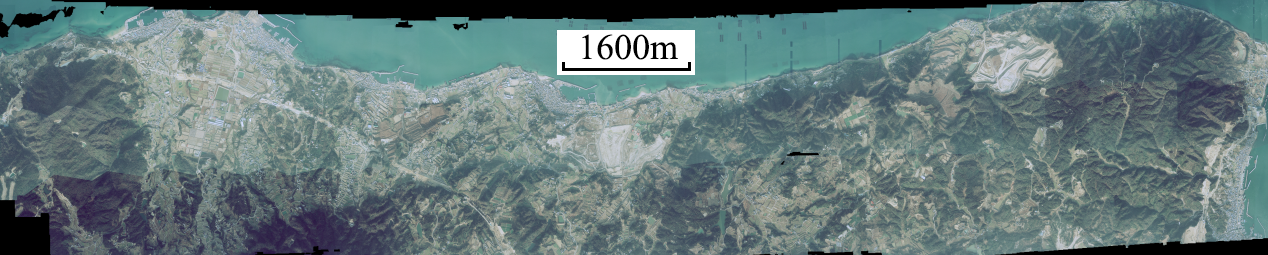}
				%\caption{DoD$_{Kobe\_1954}^{Co-Reg}$}
			\end{minipage}%
		}
		\caption{Orthophotos of the datasets used in experiments.}
		\label{DOM}
	\end{center}
\end{figure*}

\begin{figure*}[htbp]
	\begin{center}
		\subfigure[Fr{\'e}jus 1954]{
			\begin{minipage}[t]{0.48\linewidth}
				\centering
				\includegraphics[width=6.5cm, trim=100 7 100 16, clip]{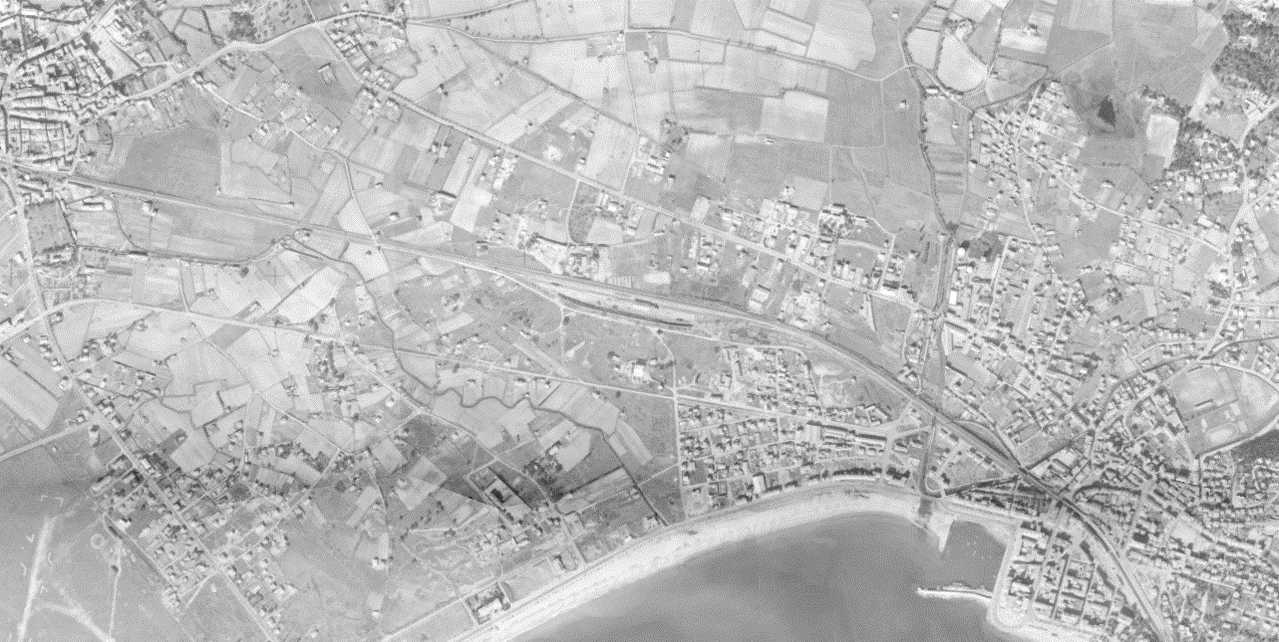}
				%\caption{DoD$_{Pezenas\_1954}^{Co-Reg}$}
			\end{minipage}%
		}
		\subfigure[Fr{\'e}jus 2014]{
			\begin{minipage}[t]{0.48\linewidth}
				\centering
				\includegraphics[width=6.5cm, trim=100 0 100 0, clip]{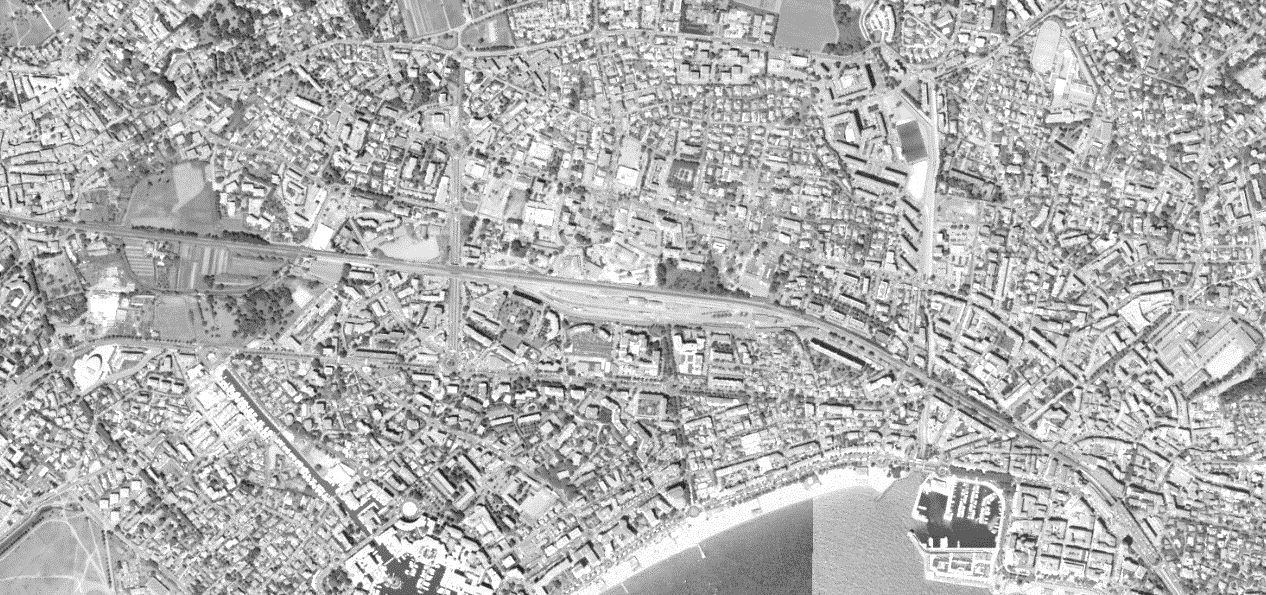}
				%\caption{DoD$_{Pezenas\_1954}^{Co-Reg}$}
			\end{minipage}%
		}
		\caption{Evolution of a subregion in Fr{\'e}jus.}
		\label{FrejusEvolution}
	\end{center}
\end{figure*}

We tested our method on three datasets: Pezenas, Fr{\'e}jus and Kobe. Details of the datasets are listed in Table~\ref{Data details} and Figure~\ref{DOM}.\\
\paragraph{Pezenas} It is a 420 $km^2$ rectangular area located in Pezenas in the Occitanie region in southern France. The area is mainly covered with vegetation and several sparsely populated urban zones. We have at our disposal three sets of images acquired in 1971, 1981 and 2015. The epoch 2015 was acquired with the IGN's digital metric camera ~\cite{souchon2010ign}, and is treated as the ground truth (GT) during our processing \MajorMod{(in other words, the 2015 image orientations are fixed during BBA)}. The area exhibits changes in scene appearance in the 44-year period.\\
\paragraph{Fr{\'e}jus} It is a 15 $km^2$ rectangular area located in Fr{\'e}jus, a commune in southeastern France. The area is mainly covered with buildings along with scattered farmlands, except a half-moon-shaped bay located in south. We have four sets of images acquired in 1954, 1966, 1970 and 2014. The epoch 2014 is treated as GT. The area exhibits drastic scene changes in the 60-year period, as can be seen in Figure~\ref{FrejusEvolution}, where the evolution of a subregion is displayed.\\
\paragraph{Kobe} It is a 90 $km^2$ area of irregular shape located in the north of Awaji Island, Japan. The well-known Kobe earthquake happened here in January 1995. We have two sets of images: pre-event acquired in 1991 and post-event acquired in 1995. It is mainly covered with mountain area and narrow urban zones along the sea. There is no GT, hence we measured 2 points on Google map to roughly scale the result to metric units. In this dataset we are interested in localizing the earthquake fault.
\begin{table}[htbp]
	\scriptsize %\footnotesize
	\centering
	\begin{tabular}{||l|c|c|c|c||c|c|c|c||c|c||}\hline
		&\multicolumn{4}{c||}{Pezenas}&\multicolumn{4}{c||}{Fr{\'e}jus}&\multicolumn{2}{c||}{Kobe}\\\hline
		&E1971&E1981&\multicolumn{2}{c||}{E2015}&E1954&E1966&E1970&E2014&E1991&E1995\\\hline\hline
F [pix]&7589&7607&9967.5&9204.5&23350&10230&10230&\color{black}18281&7662&7662\\
%Size [mm]&230230&230230&47,35&50,36&\color{black}300,300&\color{black}180,180&\color{black}180,180&99.28,72.42&212212&212,212\\
Wid [mm]&230&230&47&50&300&180&180&99.28&212&212\\
Hei [mm]&230&230&35&36&300&180&180&72.42&212&212\\
GSD [m]&0.32&0.59&0.46&0.5&\color{black}0.11&\color{black}0.17&0.17&0.35&0.5&0.18\\
F. o.&   60\%&60\%&60\%&60\%&60\%&60\%&60\%&60\%&   65\%&65\%\\
S. o.&   20\%&20\%&50\%&50\%&20\%&30\%&30\%&30\%&   35\%&65\%\\
H  [m]&2400&4500&4600&4600&2500&1700&1700&6500&3800&1400\\
Nb &57&27&308&74&19&15&19&33&15&83\\\hline
	\end{tabular}
	\caption{Dataset details of Pezenas, Fr{\'e}jus and Kobe. The 2015 acquisition of Pezenas is obtained with two sets of camera. E stands for epoch, F means focal length, Wid and Hei are the width and height of image, GSD is the ground sampling distance, F.o. and S.o. are forward and side overlap, H is the flying height, Nb is the number of images.}
	\label{Data details}
\end{table}

\subsection{Comparison of co-registration with DSMs and orthophotos}
\begin{figure*}[htbp]
	\begin{center}
		\subfigure[Overlapping area$_{Orthophoto}$]{
	\begin{minipage}[t]{0.48\linewidth}
		\centering
		\includegraphics[width=6cm]{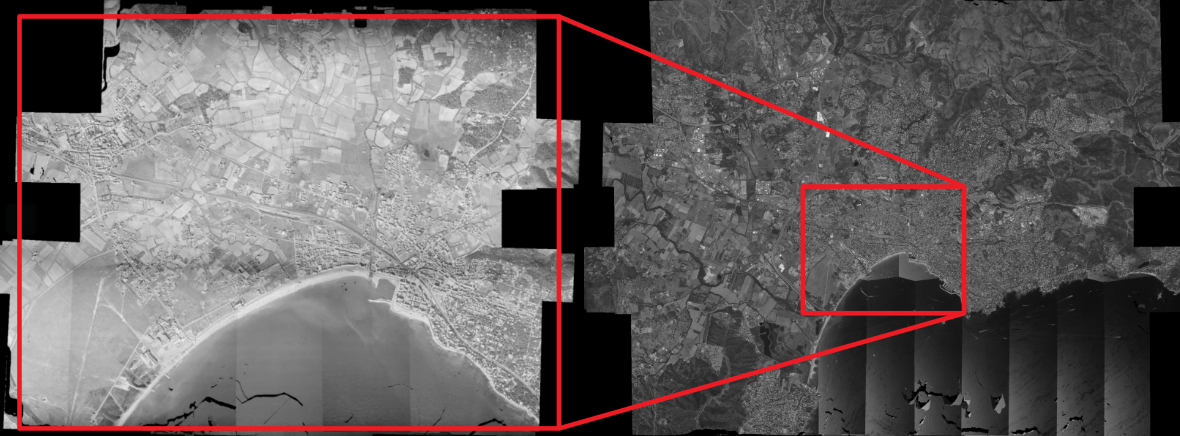}
		%\caption{DoD$_{Pezenas\_1971}^{Co-Reg}$}
	\end{minipage}%
}
\subfigure[Overlapping area$_{DSM}$]{
	\begin{minipage}[t]{0.48\linewidth}
		\centering
		\includegraphics[width=6cm]{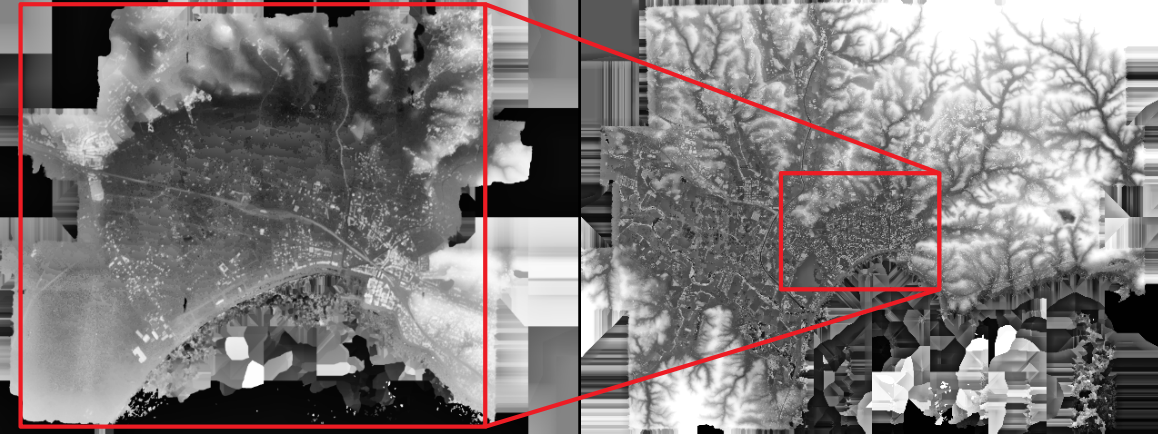}
		%\caption{DoD$_{Pezenas\_1971}^{Guided}$}
	\end{minipage}  
}  
		\subfigure[Ours$_{Orthophoto}$]{
			\begin{minipage}[t]{0.48\linewidth}
				\centering
				\includegraphics[width=6cm]{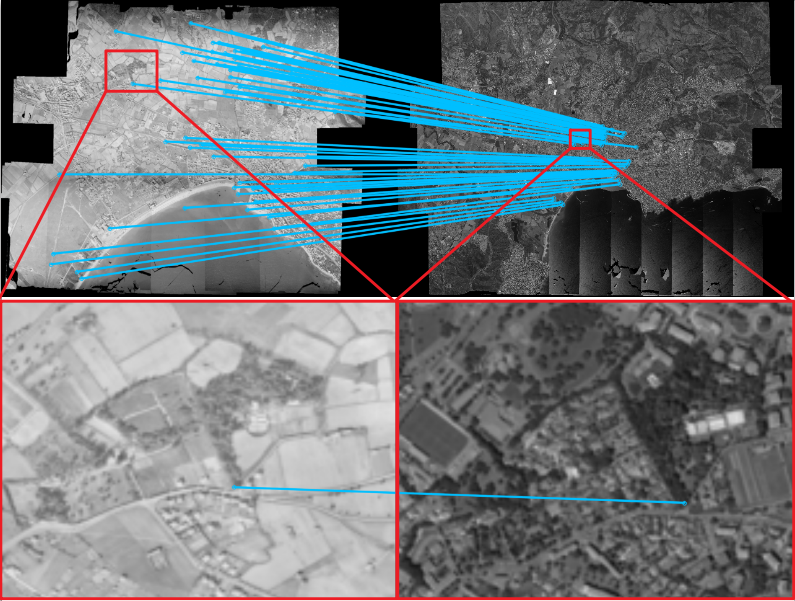}
				%\caption{DoD$_{Pezenas\_1971}^{Co-Reg}$}
			\end{minipage}%
		}
		\subfigure[Ours$_{DSM}$]{
			\begin{minipage}[t]{0.48\linewidth}
				\centering
				\includegraphics[width=6cm]{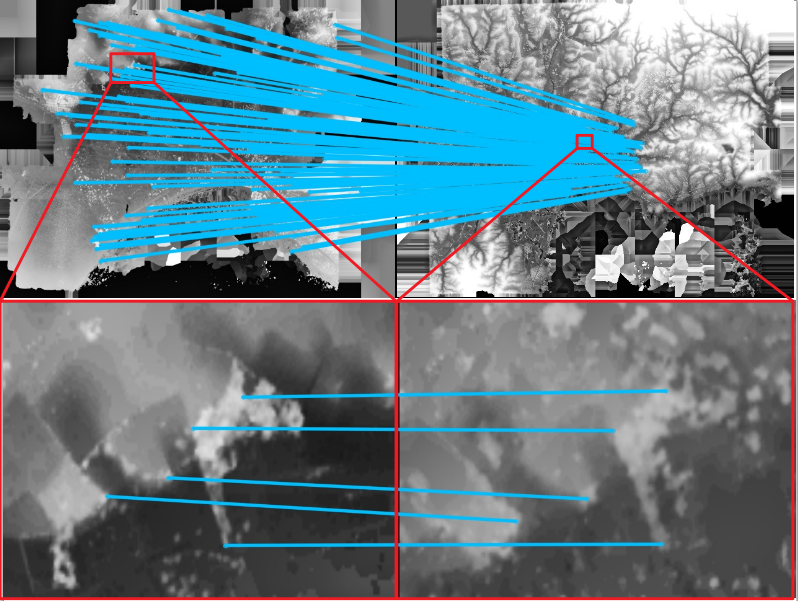}
				%\caption{DoD$_{Pezenas\_1971}^{Guided}$}
			\end{minipage}  
		}       
		\subfigure[SuperGlue$_{Orthophoto}$]{
			\begin{minipage}[t]{0.48\linewidth}
				\centering
				\includegraphics[width=6cm]{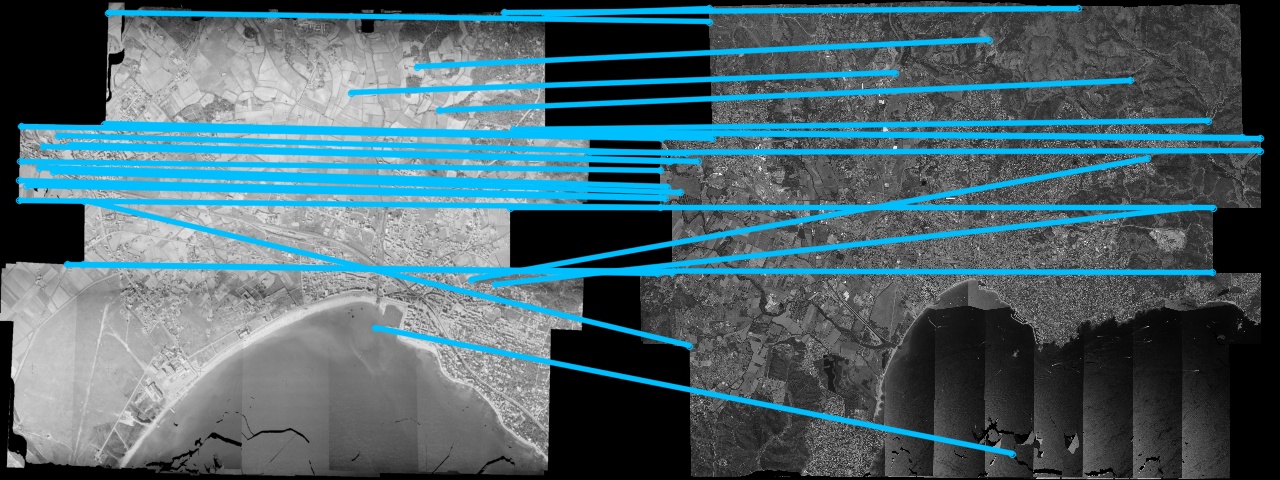}
				%\caption{DoD$_{Pezenas\_1981}^{Co-Reg}$}
			\end{minipage}%
		}
		\subfigure[SuperGlue$_{DSM}$]{
			\begin{minipage}[t]{0.48\linewidth}
				\centering
				\includegraphics[width=6cm]{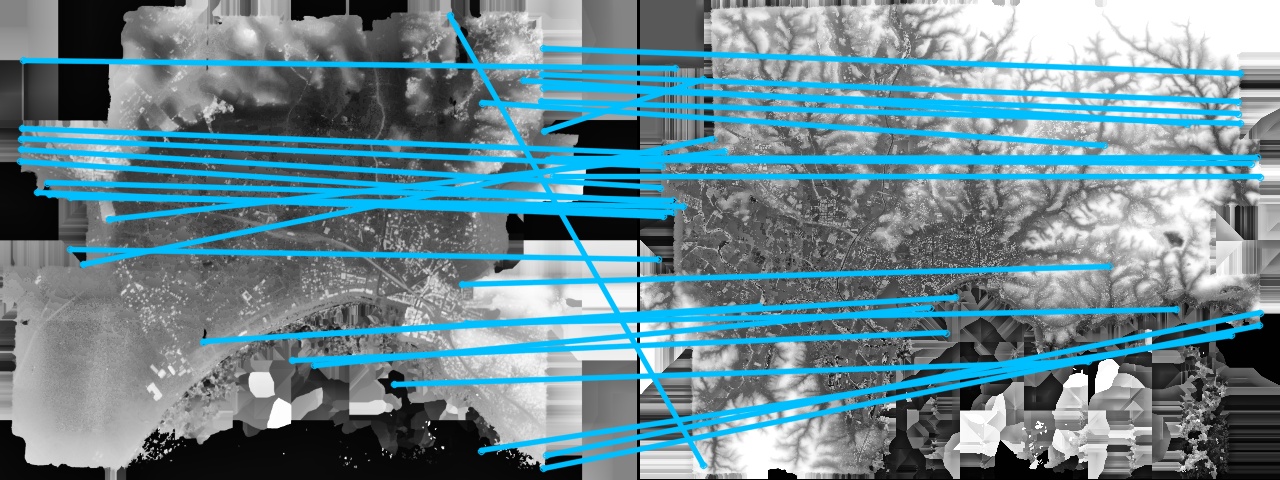}
				%\caption{DoD$_{Pezenas\_1981}^{Co-Reg}$}
			\end{minipage}%
		}
		\subfigure[SIFT$_{Orthophoto}$]{
			\begin{minipage}[t]{0.48\linewidth}
				\centering
				\includegraphics[width=6cm]{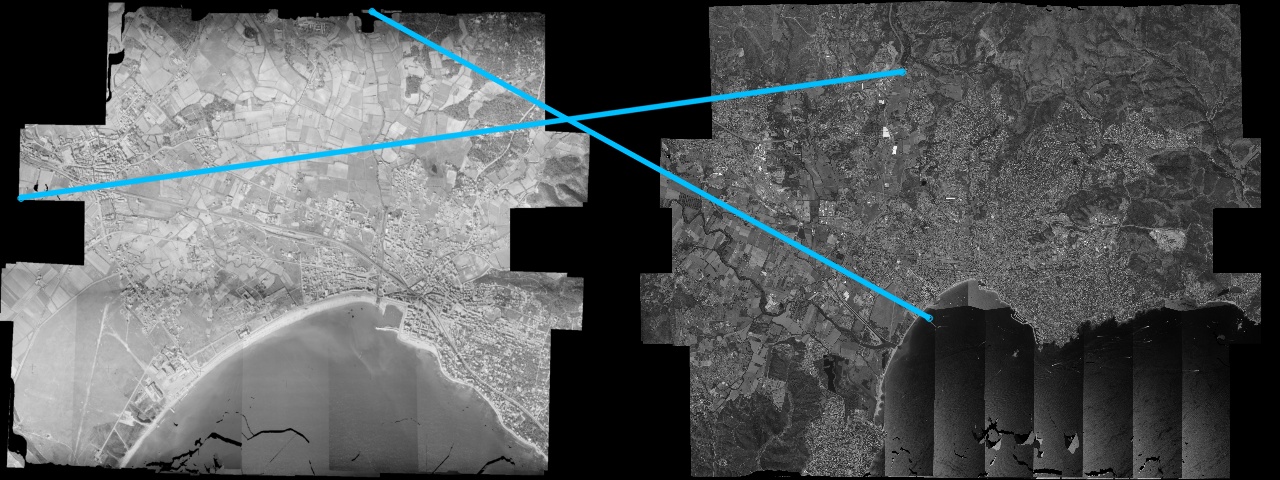}
				%\caption{DoD$_{Pezenas\_1971}^{Co-Reg}$}
			\end{minipage}%
		}
		\subfigure[SIFT$_{DSM}$]{
			\begin{minipage}[t]{0.48\linewidth}
				\centering
				\includegraphics[width=6cm]{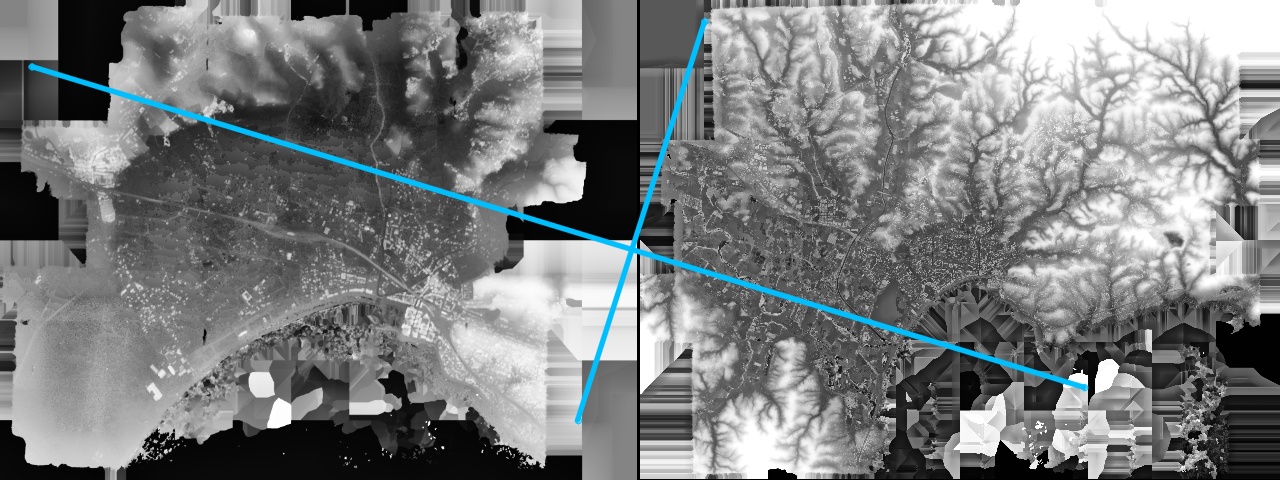}
				%\caption{DoD$_{Pezenas\_1971}^{Co-Reg}$}
			\end{minipage}%
		}
		\caption{Comparison of 3 rough co-registration methods (Ours, SuperGlue and SIFT) on 2 types of images (orthophotos and DSMs). Each subfigure displays Fr{\'e}jus 1954 on the left and 2014 on the right. \MajorMod{(a, b) The overlapping areas (red rectangles) of orthophotos and DSMs individually.} (c, e, g) Matching result of Ours, SuperGlue and SIFT on orthophotos. (d, f, h) Matching result of Ours, SuperGlue and SIFT on DSMs.}
		\label{roughmatchtest}
	\end{center}
\end{figure*}

In this section, we carry out the rough co-registration twice, with DSMs and orthophotos. We choose {Fr{\'e}jus in the years 1954 and 2014 for this experiment because it is the most difficult matching scenario, demonstrating significant scene changes} \MajorMod{and very different area extents}. \MajorMod{As can be seen in Figure~\ref{roughmatchtest}(a-b), {the whole of the epoch 1954 overlaps by solely $\approx$10 \% with the epoch 2014}.} The sizes of orthophotos for Fr{\'e}jus 1954 and 2014 individually are 519$\times$363 pixels and 1888$\times$1361 pixels. The DSM has the same size as orthophoto.
%Their corresponding image sizes are displayed in Table~\ref{DSMsize}. 
{For both types of images (i.e., DSM rasters and orthophotos)} we apply \textit{Ours} (\MajorMod{\textit{one-to-many tiling} scheme} {with} SuperGlue, followed by RANSAC), {pure} SuperGlue and SIFT. Feature correspondences are visualised in Figure~\ref{roughmatchtest} (c-h) and the inlier ratios are given in Table~\ref{roughmatchinlier}.\\
As can be seen, \textit{Ours} inlier ratio reached 18.6\% and 37.1\% for orthophotos and DSMs, respectively. SuperGlue and SIFT reached 0\% for both types of images. DSMs tend to have more inliers than orthophotos, because there is much less perceived change in the DSMs than there is in the orthophotos, see Figure~\ref{roughmatchtest} (c) and (d).
\MajorMod{Our rough co-registration is capable of recovering robust correspondences because \textit{one-to-many tiling scheme} finds many good correspondences in the overlapping tile pairs, and few wrong ones in the non-overlapping pairs, which ensures the success of RANSAC.}
%\begin{table}[htbp]
%	\centering
%	\begin{tabular}{||l|c|c||}\hline 
%		&Fr{\'e}jus 1954&Fr{\'e}jus 2014\\\hline\hline
%		orthophoto &519$\times$363 pixels&1888$\times$1361 pixels\\\hline	
%		DSM&519$\times$363 pixels&1888$\times$1361 pixels\\\hline		
%	\end{tabular}
%	\caption{Image size of orthophotos and DSMs used in Figure~\ref{roughmatchtest}. \er{you can make this table 1-row table because ortho and DSM have the same size; or actually delete the table and put the size in the text;}}
%	\label{DSMsize}
%\end{table} 

\begin{table}[htbp]
	\centering
	\begin{tabular}{||l|l|c|c|c||}\hline 
		& &Correspondences&&\\
		& &number&Inlier number&Inlier ratio\\\hline\hline
		\multirow{3}{*}{orthophoto}& SIFT&2&0&0\%\\\cline{2-5}
		&SuperGlue&29&0&0\%\\\cline{2-5}
		&Ours&290&\textbf{54}&18.6\%\\\hline		
		\multirow{3}{*}{DSM}& SIFT&2&0&0\%\\\cline{2-5}
		&SuperGlue&24&0&0\%\\\cline{2-5}
		&Ours&302&\textbf{112}&37.1\%\\\hline
	\end{tabular}
	\caption{Feature correspondences number, inlier number and inlier ratio in Figure~\ref{roughmatchtest}. 3 rough co-registration methods (Ours, SuperGlue and SIFT) on 2 types of images (orthophoto and DSM) are displayed.}
	\label{roughmatchinlier}
\end{table} 

\lz{
\subsection{Comparison of precise matching on DSMs and original images}
In order to decide which type of images (DSMs or original images) is more suitable for executing the precise matching, we apply our pipeline \textit{Patch} on both DSMs and original images of Fr{\'e}jus 1970 and 2014.
The final feature correspondences are displayed in Figure~\ref{precisematchingdepth} (a) and (b). 
To asses quantitatively the results, we created a GT depth map and calculated the accuracy (correct matches / total matches). In Figure~\ref{precisematchingdepth}~(c) we plot the accuracy curves while varying the reprojection error threshold from 0 to 10 pixels. 
{It is clear that} the result using the original images is more accurate, even though the DSMs recovered more correspondences.
This is because historical DSMs at full resolution are too noisy to guarantee high precision measurements (see the DSM shaded image in Figure~\ref{precisematchingdepth} (d).
\begin{figure*}[htbp]
	\begin{center}
		\subfigure[Feature correspondences on RGB images]{
			\begin{minipage}[t]{0.48\linewidth}
				\centering
				\includegraphics[width=6cm]{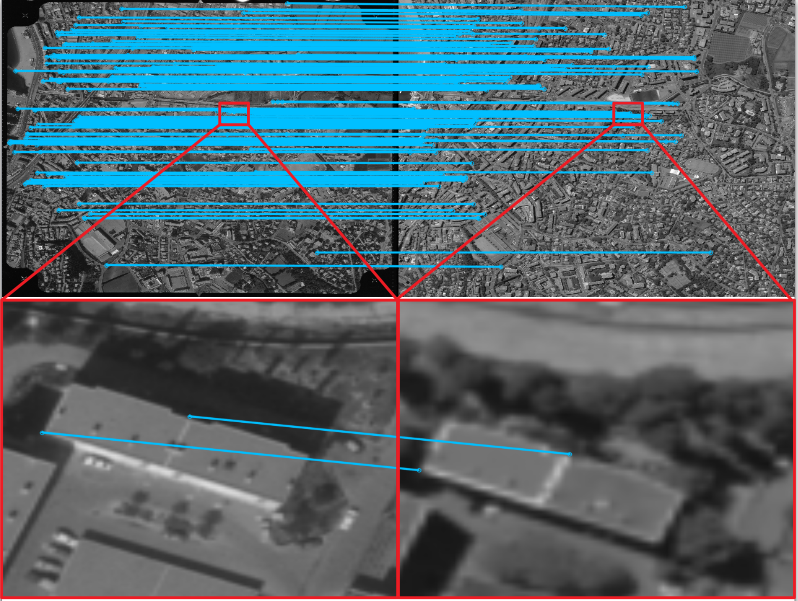}
				%\caption{DoD$_{Pezenas\_1971}^{Co-Reg}$}
			\end{minipage}%
		}
		\subfigure[Feature correspondences on DSMs]{
			\begin{minipage}[t]{0.48\linewidth}
				\centering
				\includegraphics[width=6cm]{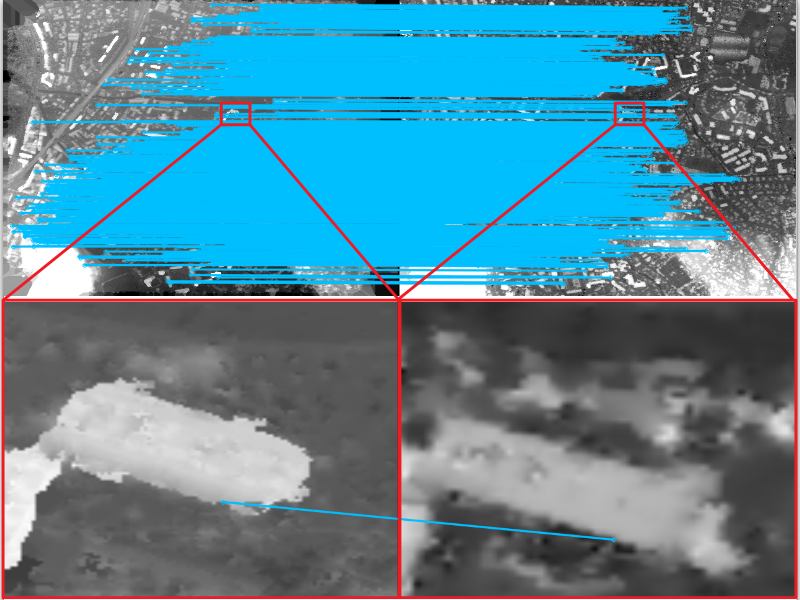}
				%\caption{DoD$_{Pezenas\_1971}^{Guided}$}
			\end{minipage}  
		}       
		\subfigure[Accuracy of (a) and (b)]{
			\begin{minipage}[t]{0.58\linewidth}
				\centering
				\includegraphics[width=7cm]{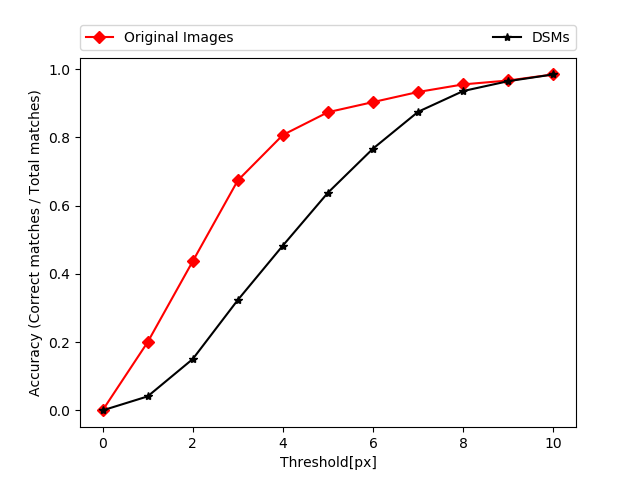}
				%\caption{DoD$_{Pezenas\_1981}^{Co-Reg}$}
			\end{minipage}%
		}
			\subfigure[shaded image of historical DSM]{
		\begin{minipage}[t]{0.38\linewidth}
			\centering
			\includegraphics[width=5cm]{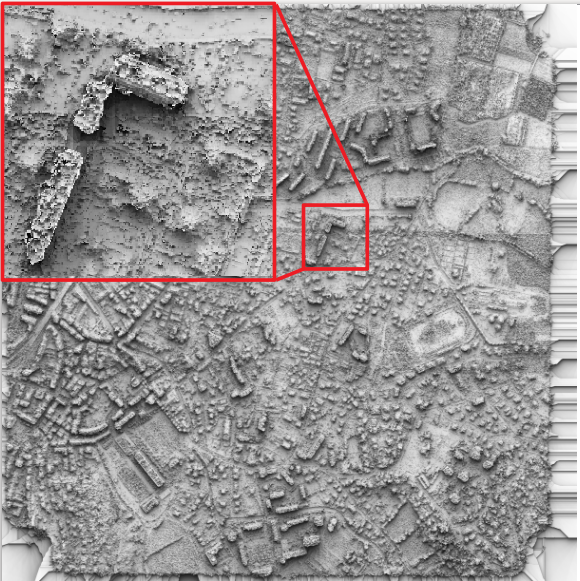}
			%\caption{DoD$_{Pezenas\_1981}^{Co-Reg}$}
		\end{minipage}%
	}
		\caption{Comparison of precise matching on original RGB images and DSMs.}
		\label{precisematchingdepth}
	\end{center}
\end{figure*} 
}
\subsection{Comparison between Ours (\textit{Guided} and \textit{Patch}), SuperGlue and SIFT}\label{comp}
\begin{figure*}[htbp]
	\begin{center}
		\subfigure[SuperGlue on images with rotation]{
			\begin{minipage}[t]{0.48\linewidth}
				\centering
				\includegraphics[width=6cm]{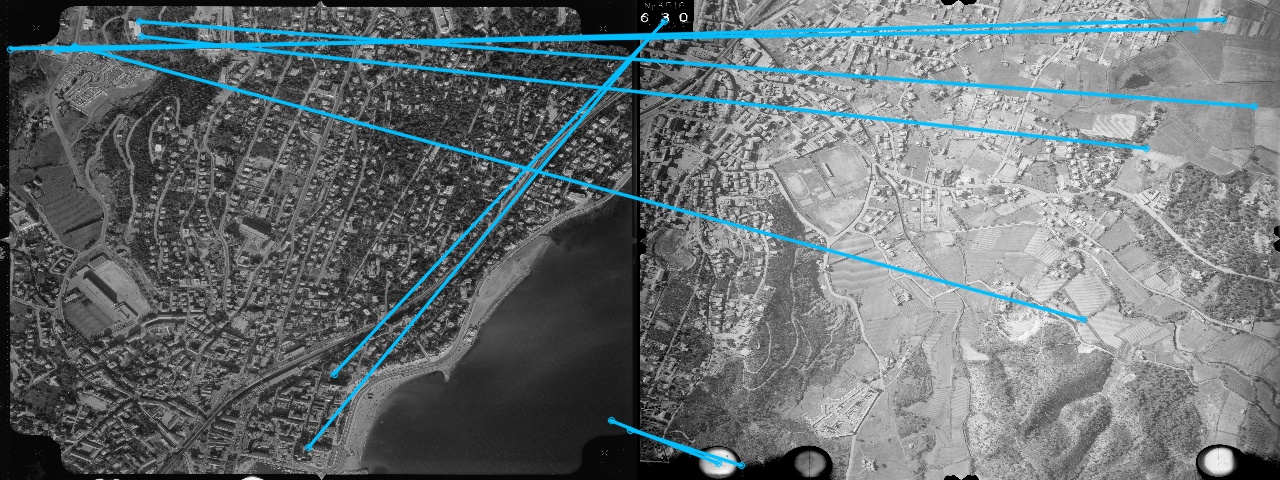}
				%\caption{DoD$_{Pezenas\_1971}^{Co-Reg}$}
			\end{minipage}%
		}
		\subfigure[\textit{Guided}]{
			\begin{minipage}[t]{0.48\linewidth}
				\centering
				\includegraphics[width=6cm]{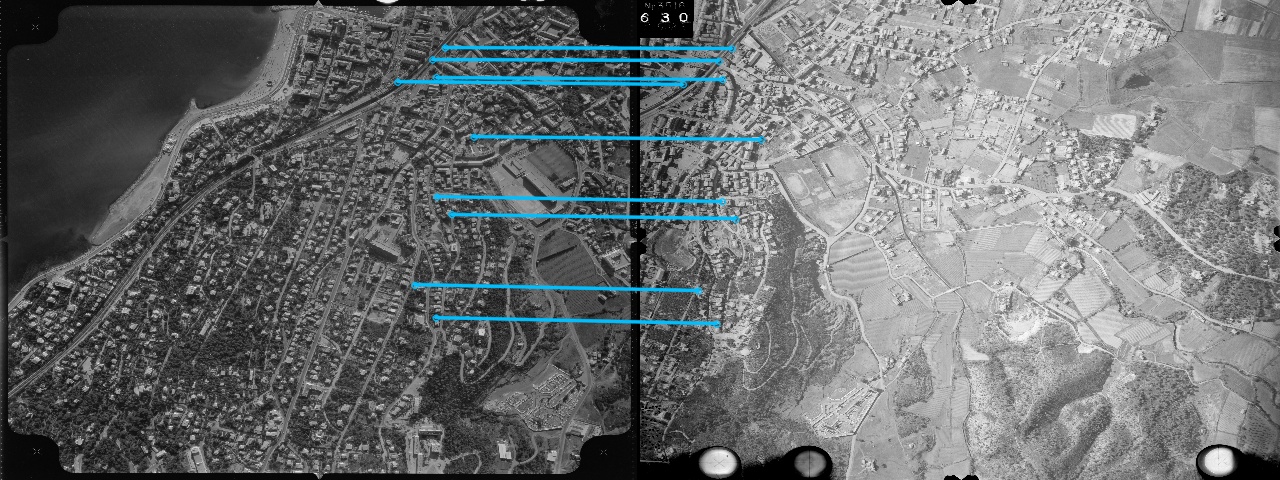}
				%\caption{DoD$_{Pezenas\_1971}^{Guided}$}
			\end{minipage}  
		}       
		\subfigure[\textit{Patch}]{
			\begin{minipage}[t]{0.48\linewidth}
				\centering
				\includegraphics[width=6cm]{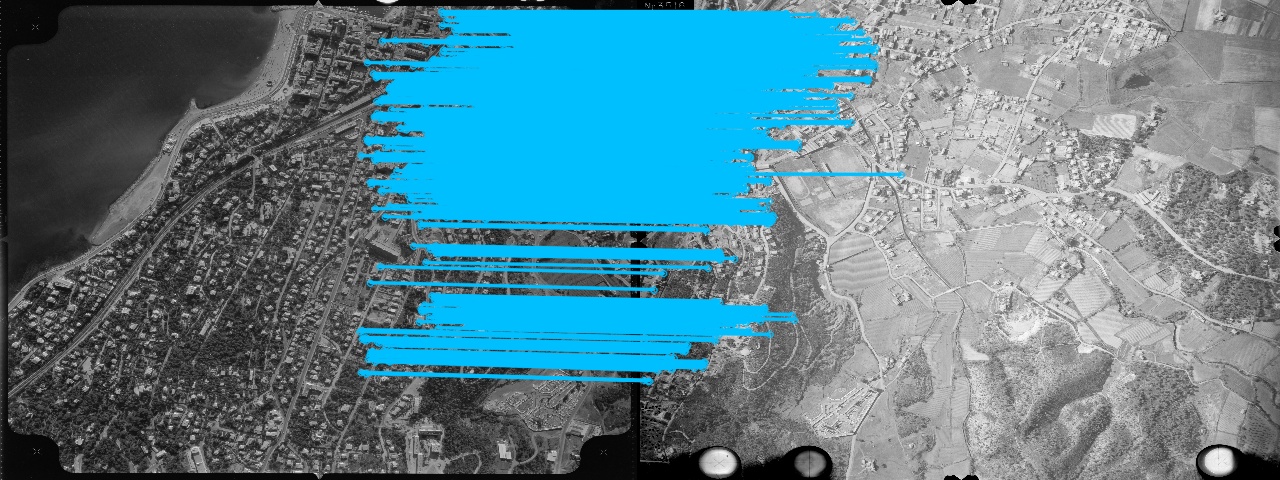}
				%\caption{DoD$_{Pezenas\_1981}^{Co-Reg}$}
			\end{minipage}%
		}
		\subfigure[SIFT]{
			\begin{minipage}[t]{0.48\linewidth}
				\centering
				\includegraphics[width=6cm]{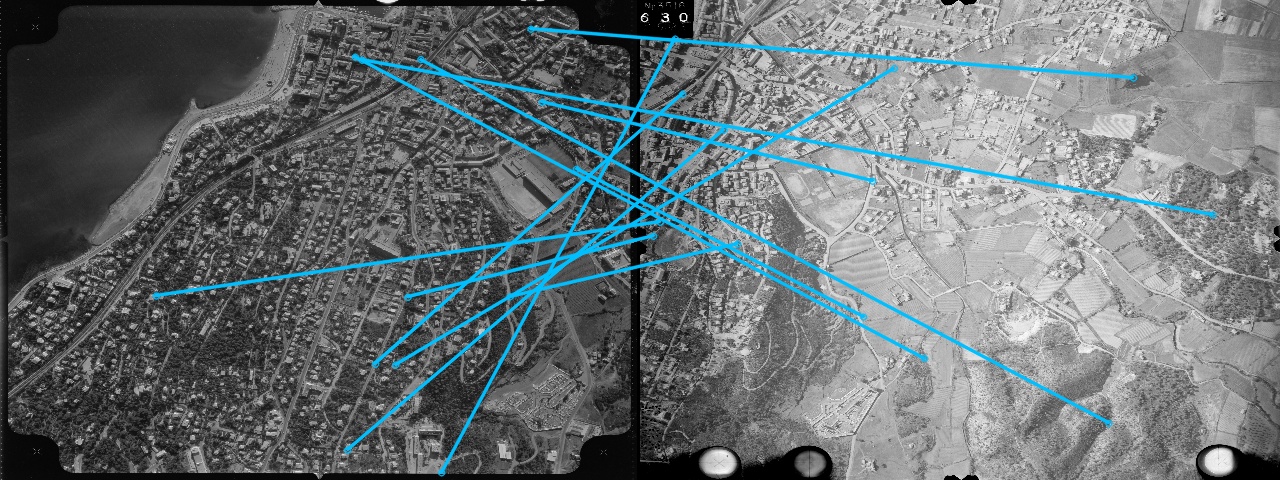}
				%\caption{DoD$_{Pezenas\_1971}^{Co-Reg}$}
			\end{minipage}%
		}
		\subfigure[SuperGlue]{
		\begin{minipage}[t]{0.48\linewidth}
			\centering
			\includegraphics[width=6cm]{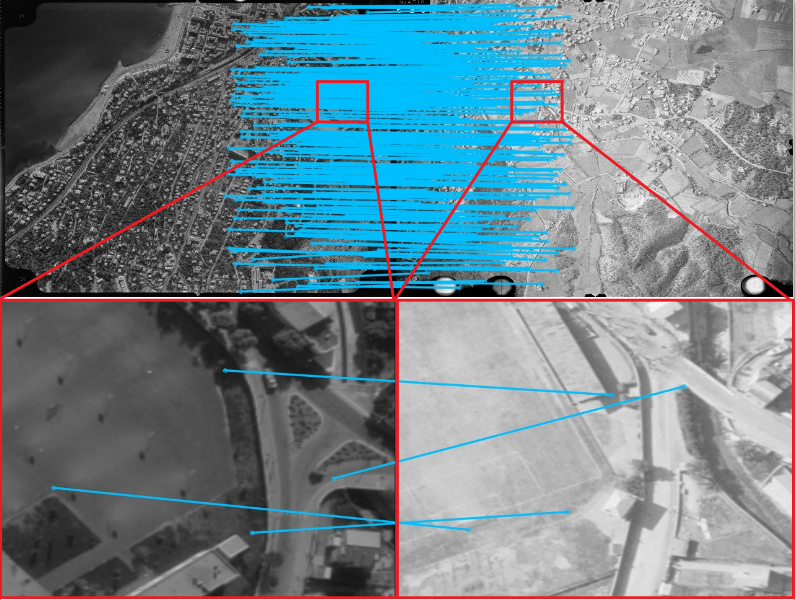}
			%\caption{DoD$_{Pezenas\_1971}^{Patch}$}
		\end{minipage}  
	}
	\subfigure[Accuracy of (b-e)]{
		\begin{minipage}[t]{0.48\linewidth}
			\centering
			\includegraphics[width=5.6cm]{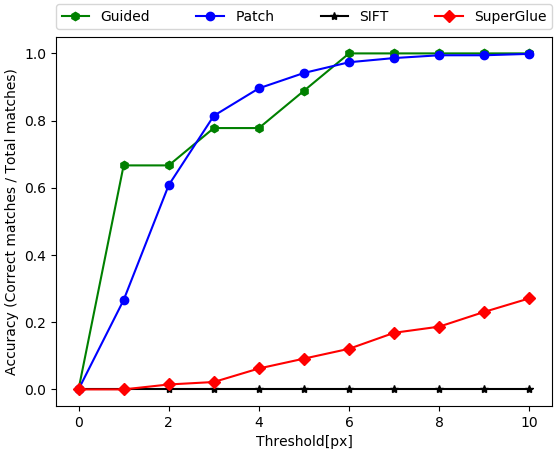}
			%\caption{DoD$_{Pezenas\_1971}^{Co-Reg}$}
		\end{minipage}%
	}
		\caption{Comparison between Ours (\textit{Guided} and \textit{Patch}), SuperGlue and SIFT. (a) Out-of-the-box SuperGlue applied on an image pair with a rotation by $180^{\circ}$. (b-e) Inter-epoch feature correspondences of our method (\textit{Guided} and \textit{Patch}), SIFT and Out-of-the-box SuperGlue applied on the same image pair without rotation. (f) Accuracy.}
		\label{comparison}
	\end{center}
\end{figure*}

In this section we compare the feature matching result between our rough-to-precise pipeline (\textit{Guided} and \textit{Patch}), SuperGlue and SIFT. The image pair used here is Fr{\'e}jus 1954 and 1966.
Given that SuperGlue failed on the original pair (cf. Figure~\ref{comparison} (a)), the left image was rotated to match the upright position of the right image prior to processing. 
SuperGlue is employed with the off-the-shelf trained outdoor model provided by the authors.
We can observe that pure SIFT fails to extract correct matches, and the out-of-the-box SuperGlue finds a lot of matches, most of which seem good, but at a closer look the details reveal poor localization precision. In \textit{Guided} and \textit{Patch}, all matches are correct, and the \textit{Patch} approach detects denser correspondences, as can be seen in Figure~\ref{comparison} (b-e).\\
In Figure~\ref{comparison}~(f) we plot the accuracy curves as we vary the reprojection error threshold from 0 to 10 pixels based on manually created GT. 
Notice that SIFT and SuperGlue had 0\% and 30\% correct matches under 10 pixels, respectively. In the meantime, both \textit{Guided} and \textit{Patch} reached over 90\% under 6 pixels.
\subsection{Evaluation Method}
\begin{figure*}[htbp]
	\begin{center}
		\subfigure[DoD$_{Pezenas\_1971}^{3vGCPs}$]{
			\begin{minipage}[t]{0.17\linewidth}
				\centering
				\includegraphics[width=2.5cm]{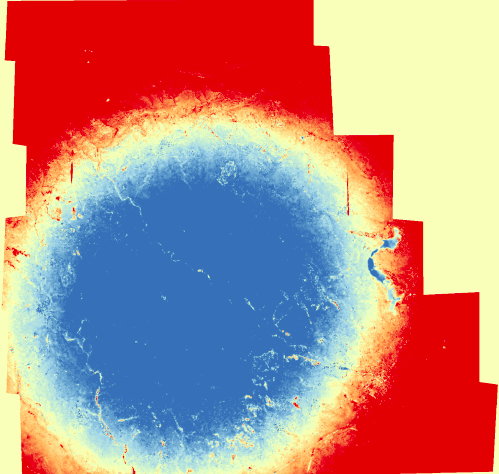}
				%\caption{DoD$_{Pezenas\_1971}^{Co-Reg}$}
			\end{minipage}%
		}
		\subfigure[DoD$_{Pezenas\_1971}^{Co-Reg}$]{
			\begin{minipage}[t]{0.17\linewidth}
				\centering
				\includegraphics[width=2.5cm]{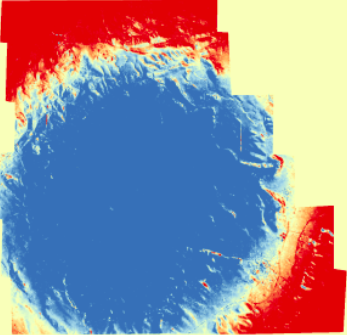}
				%\caption{DoD$_{Pezenas\_1971}^{Co-Reg}$}
			\end{minipage}%
		}
		\subfigure[DoD$_{Pezenas\_1971}^{PureSG}$]{
			\begin{minipage}[t]{0.17\linewidth}
				\centering
				\includegraphics[width=2.5cm]{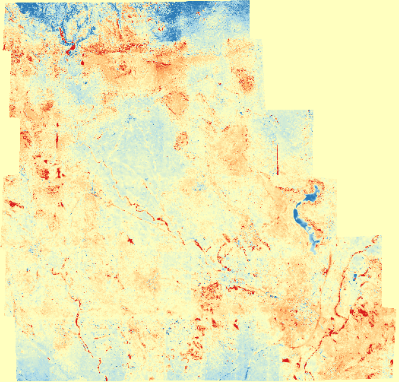}
				%\caption{DoD$_{Pezenas\_1971}^{Co-Reg}$}
			\end{minipage}%
		}
		\subfigure[DoD$_{Pezenas\_1971}^{Guided}$]{
			\begin{minipage}[t]{0.17\linewidth}
				\centering
				\includegraphics[width=2.5cm]{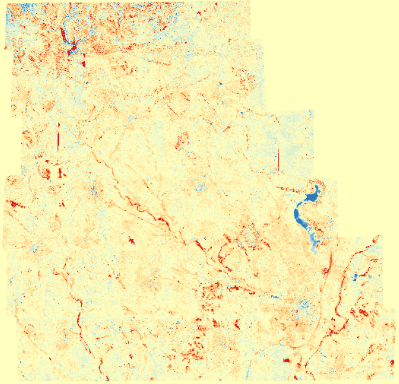}
				%\caption{DoD$_{Pezenas\_1971}^{Guided}$}
			\end{minipage}  
		}  
		\subfigure[DoD$_{Pezenas\_1971}^{Patch}$]{
			\begin{minipage}[t]{0.17\linewidth}
				\centering
				\includegraphics[width=2.5cm]{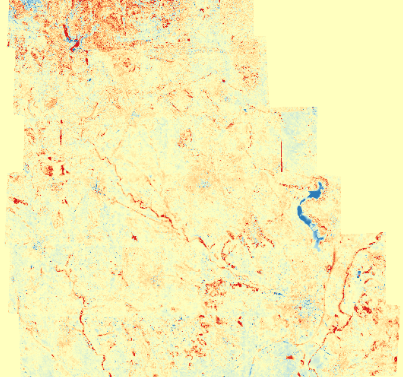}
				%\caption{DoD$_{Pezenas\_1971}^{Patch}$}
			\end{minipage}  
		}     
		
		\subfigure[DoD$_{Pezenas\_1981}^{3vGCPs}$]{
			\begin{minipage}[t]{0.17\linewidth}
				\centering
				\includegraphics[width=2.5cm]{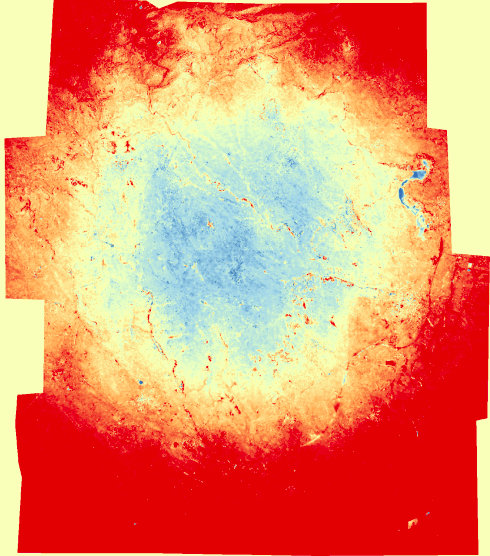}
				%\caption{DoD$_{Pezenas\_1981}^{Co-Reg}$}
			\end{minipage}%
		}        
		\subfigure[DoD$_{Pezenas\_1981}^{Co-Reg}$]{
			\begin{minipage}[t]{0.17\linewidth}
				\centering
				\includegraphics[width=2.5cm]{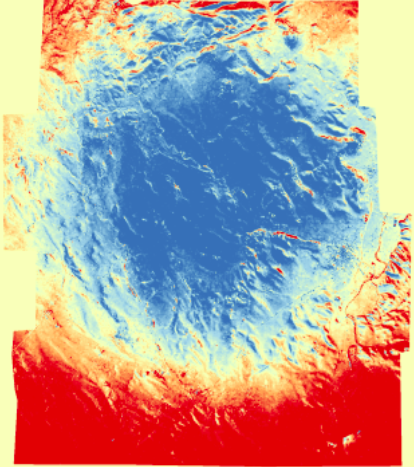}
				%\caption{DoD$_{Pezenas\_1981}^{Co-Reg}$}
			\end{minipage}%
		}
		\subfigure[DoD$_{Pezenas\_1981}^{PureSG}$]{
			\begin{minipage}[t]{0.17\linewidth}
				\centering
				\includegraphics[width=2.5cm]{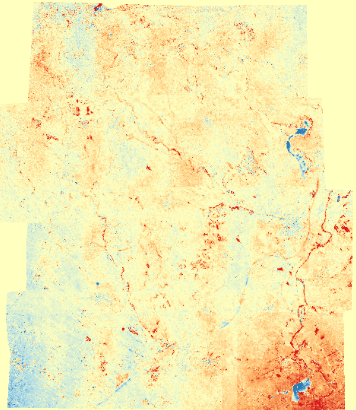}
				%\caption{DoD$_{Pezenas\_1981}^{Co-Reg}$}
			\end{minipage}%
		}
		\subfigure[DoD$_{Pezenas\_1981}^{Guided}$]{
			\begin{minipage}[t]{0.17\linewidth}
				\centering
				\includegraphics[width=2.5cm]{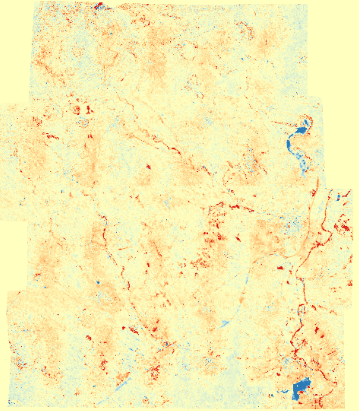}
				%\caption{DoD$_{Pezenas\_1981}^{Guided}$}
			\end{minipage}  
		}
		\subfigure[DoD$_{Pezenas\_1981}^{Patch}$]{
			\begin{minipage}[t]{0.17\linewidth}
				\centering
				\includegraphics[width=2.5cm]{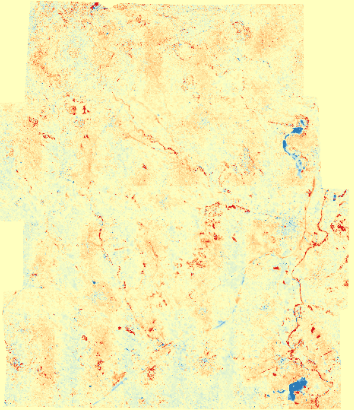}
				%\caption{DoD$_{Pezenas\_1981}^{Patch}$}
			\end{minipage}   
		}   
		\subfigure[DoD$_{Fr\acute{e}jus\_1954}^{3vGCPs}$]{
			\begin{minipage}[t]{0.17\linewidth}
				\centering
				\includegraphics[width=2.5cm]{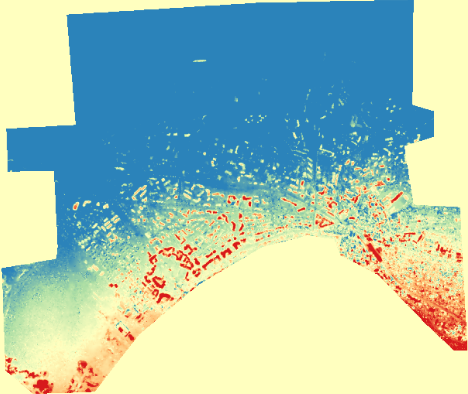}
				%\caption{DoD$_{Frejus\_1954}^{Co-Reg}$}
			\end{minipage}%
		}
		\subfigure[DoD$_{Fr\acute{e}jus\_1954}^{Co-Reg}$]{
			\begin{minipage}[t]{0.17\linewidth}
				\centering
				\includegraphics[width=2.5cm]{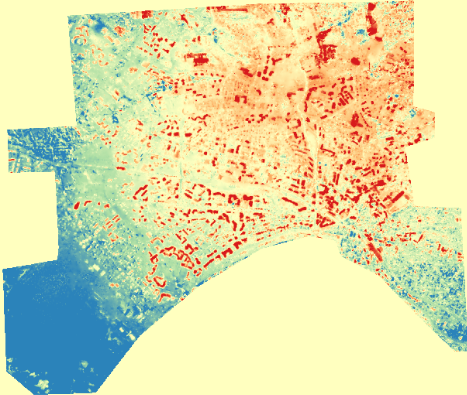}
				%\caption{DoD$_{Frejus\_1954}^{Co-Reg}$}
			\end{minipage}%
		}
		\subfigure[DoD$_{Fr\acute{e}jus\_1954}^{PureSG}$]{
			\begin{minipage}[t]{0.17\linewidth}
				\centering
				\includegraphics[width=2.5cm]{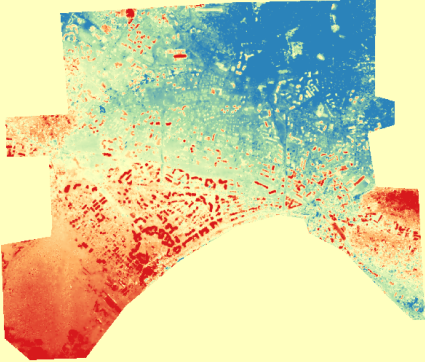}
				%\caption{DoD$_{Frejus\_1954}^{Co-Reg}$}
			\end{minipage}%
		}
		\subfigure[DoD$_{Fr\acute{e}jus\_1954}^{Guided}$]{
			\begin{minipage}[t]{0.17\linewidth}
				\centering
				\includegraphics[width=2.5cm]{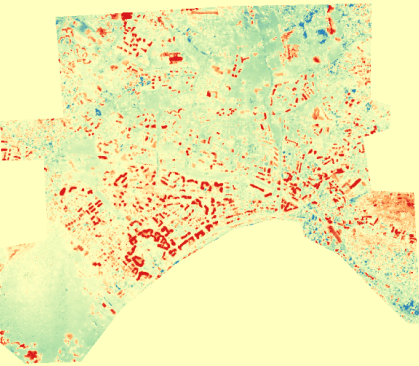}
				%\caption{DoD$_{Frejus\_1954}^{Guided}$}
			\end{minipage}  
		}      
		\subfigure[DoD$_{Fr\acute{e}jus\_1954}^{Patch}$]{
			\begin{minipage}[t]{0.17\linewidth}
				\centering
				\includegraphics[width=2.5cm]{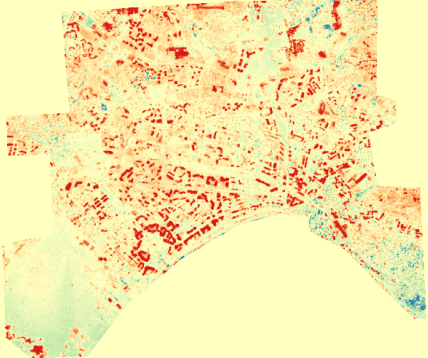}
				%\caption{DoD$_{Frejus\_1954}^{Patch}$}
			\end{minipage}  
		}
		
		\subfigure[DoD$_{Fr\acute{e}jus\_1966}^{3vGCPs}$]{
			\begin{minipage}[t]{0.17\linewidth}
				\centering
				\includegraphics[width=2.5cm]{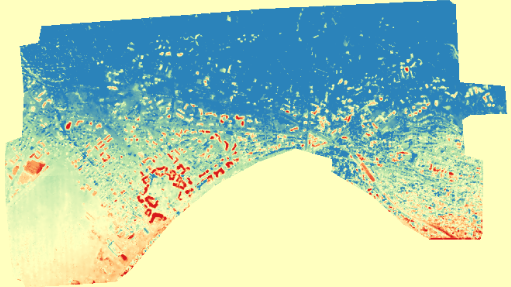}
				%\caption{DoD$_{Frejus\_1966}^{Co-Reg}$}
			\end{minipage}%
		}        
		\subfigure[DoD$_{Fr\acute{e}jus\_1966}^{Co-Reg}$]{
			\begin{minipage}[t]{0.17\linewidth}
				\centering
				\includegraphics[width=2.5cm]{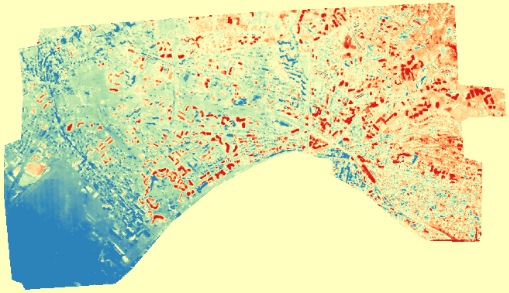}
				%\caption{DoD$_{Frejus\_1966}^{Co-Reg}$}
			\end{minipage}%
		}
		\subfigure[DoD$_{Fr\acute{e}jus\_1966}^{PureSG}$]{
			\begin{minipage}[t]{0.17\linewidth}
				\centering
				\includegraphics[width=2.5cm]{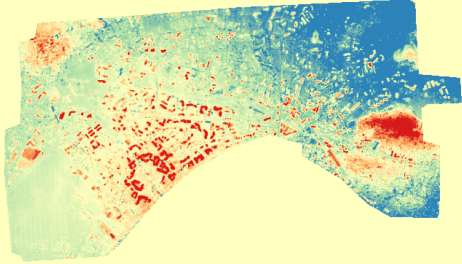}
				%\caption{DoD$_{Frejus\_1954}^{Co-Reg}$}
			\end{minipage}%
		}
		\subfigure[DoD$_{Fr\acute{e}jus\_1966}^{Guided}$]{
			\begin{minipage}[t]{0.17\linewidth}
				\centering
				\includegraphics[width=2.5cm]{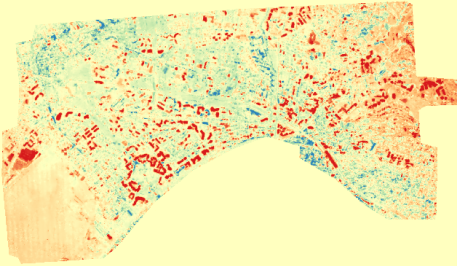}
				%\caption{DoD$_{Frejus\_1966}^{Guided}$}
			\end{minipage}  
		}  
		\subfigure[DoD$_{Fr\acute{e}jus\_1966}^{Patch}$]{
			\begin{minipage}[t]{0.17\linewidth}
				\centering
				\includegraphics[width=2.5cm]{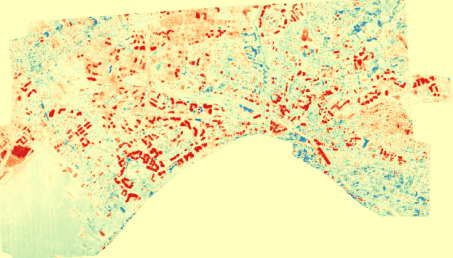}
				%\caption{DoD$_{Frejus\_1966}^{Patch}$}
			\end{minipage}   
		}
		
		\subfigure[DoD$_{Fr\acute{e}jus\_1970}^{3vGCPs}$]{
			\begin{minipage}[t]{0.17\linewidth}
				\centering
				\includegraphics[width=2.5cm]{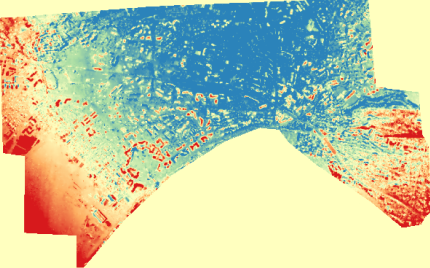}
				%\caption{DoD$_{Frejus\_1970}^{Co-Reg}$}
			\end{minipage}%
		}        
		\subfigure[DoD$_{Fr\acute{e}jus\_1970}^{Co-Reg}$]{
			\begin{minipage}[t]{0.17\linewidth}
				\centering
				\includegraphics[width=2.5cm]{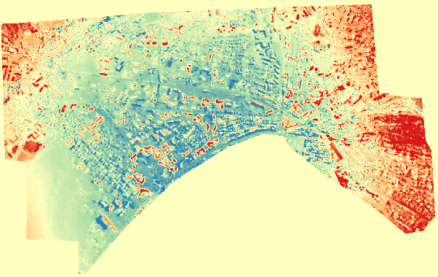}
				%\caption{DoD$_{Frejus\_1970}^{Co-Reg}$}
			\end{minipage}%
		}
		\subfigure[DoD$_{Fr\acute{e}jus\_1970}^{PureSG}$]{
			\begin{minipage}[t]{0.17\linewidth}
				\centering
				\includegraphics[width=2.5cm]{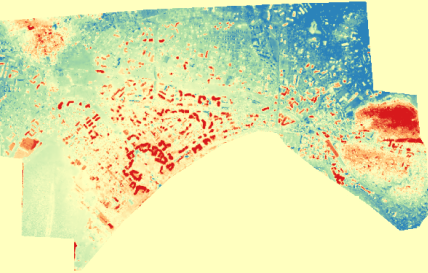}
				%\caption{DoD$_{Frejus\_1954}^{Co-Reg}$}
			\end{minipage}%
		}
		\subfigure[DoD$_{Fr\acute{e}jus\_1970}^{Guided}$]{
			\begin{minipage}[t]{0.17\linewidth}
				\centering
				\includegraphics[width=2.5cm]{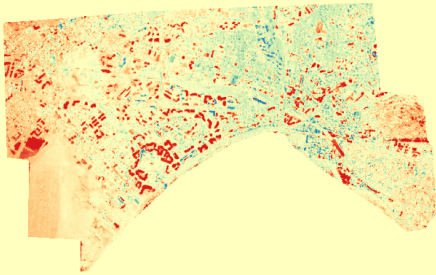}
				%\caption{DoD$_{Frejus\_1970}^{Guided}$}
			\end{minipage}  
		}   
		\subfigure[DoD$_{Fr\acute{e}jus\_1970}^{Patch}$]{
			\begin{minipage}[t]{0.17\linewidth}
				\centering
				\includegraphics[width=2.5cm]{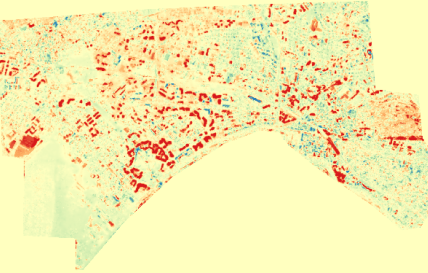}
				%\caption{DoD$_{Frejus\_1970}^{Patch}$}
			\end{minipage}    
		}
			\subfigure[DoD Legend]{
		\begin{minipage}[t]{1\linewidth}
			\centering
			\includegraphics[width=8cm]{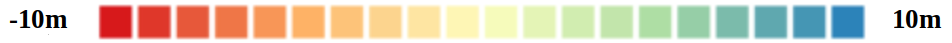}
			%\caption{DoD$_{Frejus\_1970}^{Patch}$}
		\end{minipage}    
	}
		\caption{DoDs of Pezenas and Fr{\'e}jus on 5 methods (\textit{3vGCPs}, \textit{Co-Reg}, \textit{PureSG}, \textit{Guided} and \textit{Patch})}
		\label{DoDPF}
	\end{center}
\end{figure*}

\begin{figure*}[htbp]
	\begin{center}
		\subfigure[DoD$_{Kobe}^{3vGCPs}$]{
			\begin{minipage}[t]{0.17\linewidth}
				\centering
				\includegraphics[width=8.8cm,angle=90]{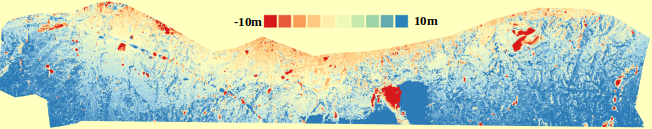}
				%\caption{DoD$_{Kobe}^{Co-Reg}$}
			\end{minipage}%
		}
		\subfigure[DoD$_{Kobe}^{Co-Reg}$]{
			\begin{minipage}[t]{0.17\linewidth}
				\centering
				\includegraphics[width=8.8cm,angle=90]{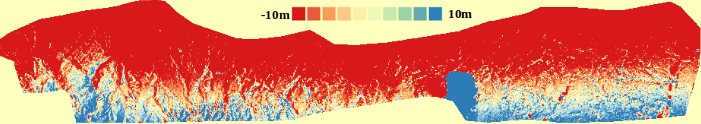}
				%\caption{DoD$_{Kobe}^{Co-Reg}$}
			\end{minipage}%
		}
		\subfigure[DoD$_{Kobe}^{PureSG}$]{
			\begin{minipage}[t]{0.17\linewidth}
				\centering
				\includegraphics[width=8.8cm,angle=90]{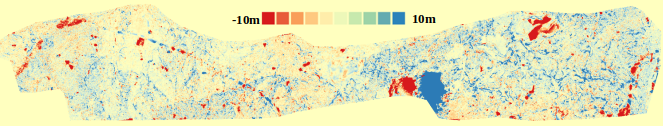}
				%\caption{DoD$_{Kobe}^{Co-Reg}$}
			\end{minipage}%
		}
		\subfigure[DoD$_{Kobe}^{Guided}$]{
			\begin{minipage}[t]{0.17\linewidth}
				\centering
				\includegraphics[width=8.8cm,angle=90]{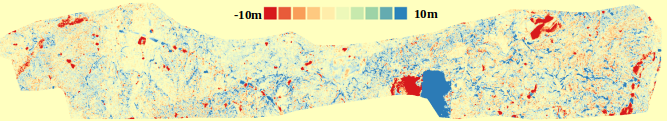}
				%\caption{DoD$_{Kobe}^{Guided}$}
			\end{minipage}  
		}       
		\subfigure[DoD$_{Kobe}^{Patch}$]{
			\begin{minipage}[t]{0.17\linewidth}
				\centering
				\includegraphics[width=8.8cm,angle=90]{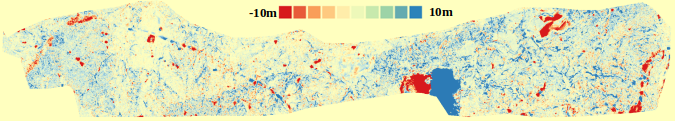}
				%\caption{DoD$_{Kobe}^{Patch}$}
			\end{minipage}  
		}
		
		\subfigure[Gd$_{Kobe}^{3vGCPs}$]{
			\begin{minipage}[t]{0.15\linewidth}
				\centering
				\includegraphics[width=8.8cm,angle=90]{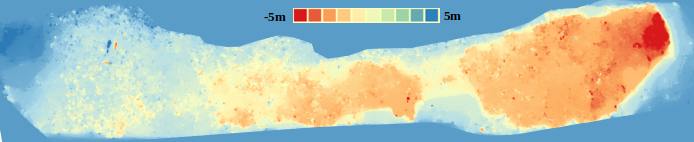}
				%\caption{Gd$_{Kobe}^{Co-Reg}$}
			\end{minipage}%
		}        
		\subfigure[Gd$_{Kobe}^{Co-Reg}$]{
			\begin{minipage}[t]{0.15\linewidth}
				\centering
				\includegraphics[width=8.8cm,angle=90]{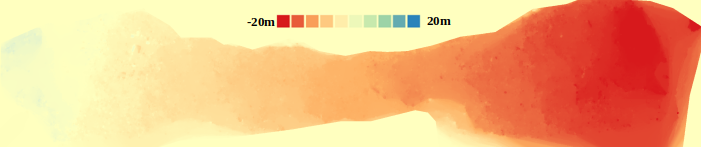}
				%\caption{Gd$_{Kobe}^{Co-Reg}$}
			\end{minipage}%
		}
		\subfigure[Gd$_{Kobe}^{PureSG}$]{
			\begin{minipage}[t]{0.15\linewidth}
				\centering
				\includegraphics[width=8.8cm,angle=90]{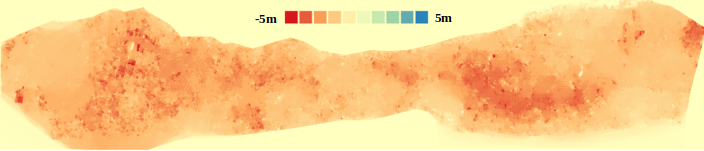}
				%\caption{Gd$_{Kobe}^{Guided}$}
			\end{minipage}  
		}       
		\subfigure[Gd$_{Kobe}^{Guided}$]{
			\begin{minipage}[t]{0.15\linewidth}
				\centering
				\includegraphics[width=8.8cm,angle=90]{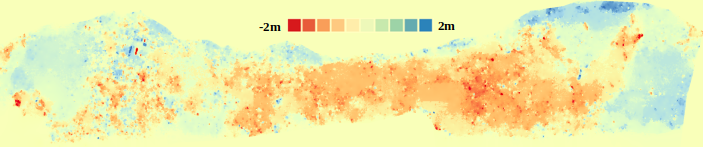}
				%\caption{Gd$_{Kobe}^{Guided}$}
			\end{minipage}  
		}       
		\subfigure[Gd$_{Kobe}^{Patch}$]{
			\begin{minipage}[t]{0.15\linewidth}
				\centering
				\includegraphics[width=8.8cm,angle=90]{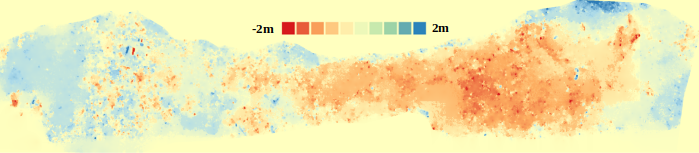}
				%\caption{Gd$_{Kobe}^{Patch}$}
			\end{minipage}      
		}       
		\subfigure[Rupture map]{
			\begin{minipage}[t]{0.1\linewidth}
				\centering
				\includegraphics[width=8.8cm,angle=90]{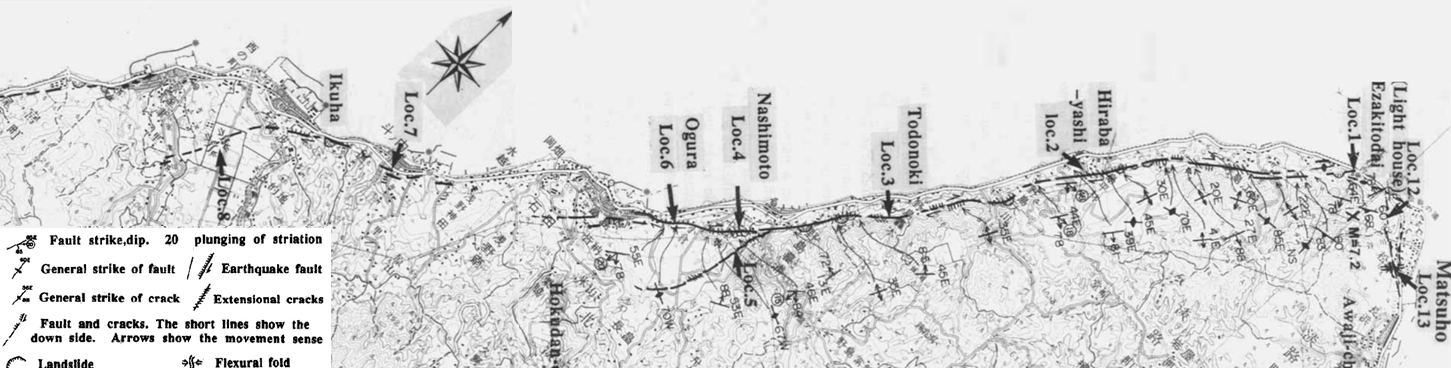}
				%\caption{Dy$_{Kobe}^{Patch}$}
			\end{minipage}  
		}
		\caption{(a-e) DoDs of Kobe on 5 methods (\textit{3vGCPs}, \textit{Co-Reg}, \textit{PureSG}, \textit{Guided} and \textit{Patch}). (f-j) Ground displacements (i.e., northeastward, denoted as $Gd$) of Kobe on 5 methods. (k) Map of earthquake surface ruptures from independent data sources.}
		\label{DoDK}
	\end{center}
\end{figure*}

We refine the co-registered orientations in a BBA routine using the inter- and intra-epoch feature correspondences (cf. Figure~\ref{Flow-process diagram}(a)). The orientations of the latest epochs in Pezenas and Fr{\'e}jus were treated as fixed during the combined BBA since they were accurately known \textit{a-priori}, while all the remaining orientations were considered as free parameters. At first, {interior orientation parameters} were shared among all images. Once stable initial values were known, interior parameters were further refined with image-dependent affine parameters. The affine component of the camera calibration is expected to model, at least partially, the shear of the analog film.\\
For comparison, we recover the image orientations with 5 methods, which can be divided into 2 categories of "rough" and "refined":
\begin{enumerate}
\item \textit{3vGCPs} ("rough"): We co-register all the epochs with 3 manually measured \textit{virtual GCPs} in total. By \textit{virtual GCPs} we mean points of known 3D coordinates in one of the epochs.
\item \textit{Co-Reg} ("rough"): We match DSMs based on \MajorMod{\textit{one-to-many tiling scheme}} to roughly co-register all the epochs (cf. Figure~\ref{Flow-process diagram}(b)).
\item \textit{PureSG} ("refined"): Based on the \textit{Co-Reg} result, we remove the scale, rotation and \MajorMod{extent} differences for inter-epoch image pairs (\MajorMod{without \textit{one-to-one tiling} scheme}), and apply off-the-shell SuperGlue to get inter-epoch feature correspondences, which are then used to refine the \textit{Co-Reg} result.
\item \textit{Guided} ("refined"): Based on the \textit{Co-Reg} result, inter-epoch feature correspondences are extracted with guided matching, as presented in Section 3.2.2. Finally, a BA routine refines the initial \textit{Co-Reg} orientations.
\item \textit{Patch} ("refined"): Same as in \textit{Guided}, except here we adopt the alternative \textit{Patch} approach as presented in Section 3.2.2.
\end{enumerate}
Our first metric to evaluate the aforementioned 5 methods is DoD ({\textit{Difference of DSMs}}). For Pezenas and Fr\'ejus datasets, DoDs are calculated between historical epochs and the available GT epochs (cf. Figure~\ref{DoDPF}). For Kobe dataset there is no GT so we calculate the DoD between 1991 and 1995 instead (cf. Figure~\ref{DoDK} (a-e)). Statistics of the presented DoDs are demonstrated in Table~\ref{Statistical information of Z coordinate in DoD}.\\
Even though DoD provides a global evaluation over the entire scene, it is contaminated with real scene changes. Therefore, we present a supplementary metric for each dataset:
\begin{enumerate}
\item For Pezenas and Fr\'ejus, to evaluate the orientations resulting from the 5 methods, we manually measured 15 check points that are uniformly distributed across images. \MajorMod{The ground coordinate of the check points are measured in the GT epoch (with measurement accuracy of 0.17m in Frejus and 0.55m in Pezenas).} Average value $\mu$, standard deviation $\sigma$, and absolute average value $|\mu|$ of the x-, y- and z-coordinate are displayed in Table~\ref{CheckptAcuracy}.
\item For Kobe, we: (1) calculated the DSMs; (2) orthorectified the images; and (3) performed 2D correlation of the respective orthophotos ~\cite{rosu2015measurement} to see whether we can observe the slip of the tectonic plate. The displacement maps (i.e., northeastward, denoted as $Gd$) are presented in Figure~\ref{DoDK}(f-j).
\end{enumerate}
\begin{table}%[H]
	\footnotesize
	\centering
	\begin{tabular}{||l|c|c|c||l|c|c|c||}\hline
		&$\mu$ [m]&$\sigma$ [m]&$|\mu|$ [m]&   &     $\mu$ [m]&$\sigma$ [m]&$|\mu|$ [m]\\\hline\hline
		DoD$_{Frejus\_1954}^{3vGCPs}$ & 12.89 & 12.35 & 14.12&     DoD$_{Pezenas\_1971}^{3vGCPs}$ &-3.89&18.07&14.06\\
		DoD$_{Frejus\_1954}^{Co-Reg}$ & 1.47 & 6.04 & 4.66&      DoD$_{Pezenas\_1971}^{Co-Reg}$ &7.33 & 17.22 & 16.09\\
		DoD$_{Frejus\_1954}^{PureSG}$ & 2.15 & 9.09 & 7.37&      DoD$_{Pezenas\_1971}^{PureSG}$ &-0.03 & 2.92 & 1.94\\
		DoD$_{Frejus\_1954}^{Guided}$ & 0.39 & 3.48 & 2.53&      DoD$_{Pezenas\_1971}^{Guided}$ &-0.48 & 2.25 & \textbf{1.38}\\
		DoD$_{Frejus\_1954}^{Patch}$ & -1.69 & 3.44 & \textbf{2.51}&       DoD$_{Pezenas\_1971}^{Patch}$ &-0.42 & 2.30 & 1.40\\\hline 
		
		DoD$_{Frejus\_1966}^{3vGCPs}$ & 7.49 & 6.93 & 8.32&     DoD$_{Pezenas\_1981}^{3vGCPs}$ &-6.80&8.89&8.21\\
		DoD$_{Frejus\_1966}^{Co-Reg}$ & 2.06 & 5.41 & 4.49&      DoD$_{Pezenas\_1981}^{Co-Reg}$ &0.46 & 9.87 & 7.97\\
		DoD$_{Frejus\_1966}^{PureSG}$ & 2.88 & 5.76 & 4.81&      DoD$_{Pezenas\_1981}^{PureSG}$ &-0.52 & 2.45 & 1.61\\
		DoD$_{Frejus\_1966}^{Guided}$ & -0.32 & 3.69 & 2.62&      DoD$_{Pezenas\_1981}^{Guided}$ &-0.45 & 2.07 & 1.29\\   
		DoD$_{Frejus\_1966}^{Patch}$ & -0.40 & 3.62 & \textbf{2.43}&       DoD$_{Pezenas\_1981}^{Patch}$ &-0.34 & 2.04 & \textbf{1.25}\\\hline 
		
		DoD$_{Frejus\_1970}^{3vGCPs}$ & 3.65 & 7.12 & 6.83&     DoD$_{Kobe}^{3vGCPs}$ &5.58 & 12.92 & 7.18\\
		DoD$_{Frejus\_1970}^{Co-Reg}$ & 1.30 & 5.34 & 4.44&       DoD$_{Kobe}^{Co-Reg}$ &-11.28 & 17.29 & 15.81\\
		DoD$_{Frejus\_1970}^{PureSG}$ & 1.99 & 4.98 & 4.17&       DoD$_{Kobe}^{PureSG}$ &2.48 & 12.11 & 4.34\\
		DoD$_{Frejus\_1970}^{Guided}$ & -0.61 & 3.76 & 2.77&        DoD$_{Kobe}^{Guided}$ & 1.79 & 12.21 & 4.13\\
		DoD$_{Frejus\_1970}^{Patch}$ & -0.74 & 3.36 & \textbf{2.17}&        DoD$_{Kobe}^{Patch}$ &1.58 & 9.18 & \textbf{3.62}\\\hline 
	\end{tabular}
	\caption{Average value $\mu$, standard deviation $\sigma$, and absolute average value $|\mu|$ of all the DoDs in Figure~\ref{DoDPF} and ~\ref{DoDK}(a-e).}
	\label{Statistical information of Z coordinate in DoD}
\end{table}

\begin{table}%[H]
	\footnotesize
	\centering
	\begin{tabular}{||l|c|c|c|c|c|c|c|c|c||}\hline
		&\multicolumn{3}{c|}{$\mu$ [m]}&\multicolumn{3}{c|}{$\sigma$ [m]}&\multicolumn{3}{c|}{$|\mu|$ [m]}\\\hline
		&x&y&z&x&y&z&x&y&z\\\hline\hline
		DoCheckPt$_{Frejus}^{3vGCPs}$ & -0.38 & 2.32 & -5.80 & 12.05 & 9.75 & 6.67 & 8.85 & 7.93 & 7.34\\
		DoCheckPt$_{Frejus}^{Co-Reg}$ & 7.23 & 6.18 & -3.08 & 13.75 & 6.98 & 3.79 & 10.60 & 7.43 & 3.96\\
		DoCheckPt$_{Frejus}^{PureSG}$ & 1.22 & 0.32 & -1.93 & 2.16 & 1.58 & 4.00 & 1.37 & 1.00 & 3.27\\
		DoCheckPt$_{Frejus}^{Guided}$ & 1.40 & 0.71 & -0.70 & 2.08 & 1.71 & 3.04 & 1.52 & 1.06 & 2.70\\
		DoCheckPt$_{Frejus}^{Patch}$ & 1.22 & 0.59 & -0.24 & 2.15 & 1.67 & 1.63 & 1.39 & 0.95 & \textbf{1.17}\\\hline
		DoCheckPt$_{Pezenas}^{3vGCPs}$ & 5.17 & -2.85 & 8.26 & 5.24 & 3.67 & 13.78 & 6.01 & 3.47 & 10.44\\
		DoCheckPt$_{Pezenas}^{Co-Reg}$ & 33.15 & 22.42 & 1.40 & 18.80 & 16.65 & 14.48 & 33.15 & 23.02 & 11.94\\
		DoCheckPt$_{Pezenas}^{PureSG}$ & -0.44 & -0.56 & 0.55 & 2.73 & 1.60 & 3.02 & 1.95 & 1.44 & 2.15\\
		DoCheckPt$_{Pezenas}^{Guided}$ & 0.17 & -0.29 & 0.31 & 2.40 & 1.42 & 1.55 & 1.57 & 1.24 & 1.22\\
		DoCheckPt$_{Pezenas}^{Patch}$ & -0.04 & -0.29 & 0.46 & 2.35 & 1.42 & 1.46 & 1.64 & 1.22 & \textbf{1.10}\\\hline
	\end{tabular}
	\caption{Accuracy of 5 different sets of orientation resulting from 5 methods (\textit{3vGCPs}, \textit{Co-Reg}, \textit{PureSG}, \textit{Guided} and \textit{Patch}), evaluated on 15 check points uniformly distributed in the block for both Fr{\'e}jus and Pezenas. Average value $\mu$, standard deviation $\sigma$, and absolute average value $|\mu|$ of the x-, y- and z-coordinate are displayed for each method.}
	\label{CheckptAcuracy}
\end{table}

\subsection{Result}

\paragraph{DoDs}
Table~\ref{Statistical information of Z coordinate in DoD} reports the average value and standard deviations computed on all the DoDs in Figure~\ref{DoDPF} and ~\ref{DoDK}(a-e).  
A \textit{dome} artifact is present in DoD$^{3vGCPs}$ and DoD$^{Co-Reg}$ for all the datasets. This kind of systematic error is known to originate from poorly modeled camera internal parameters ~\cite{giordano2018toward}. The results of DoD$^{PureSG}$ for datasets with smaller time span (e.g. Kobe and Pezenas) are relatively good, thanks to our roughly co-registered orientations and DSMs. However, the results of Fr{\'e}jus are not satisfactory -- \MajorMod{without} adopting our pipeline (i.e., patch matching, 3D-RANSAC filter and cross-correlation), the inlier ratio and precision of the inter-epoch feature correspondences are low.
DoD$^{Patch}$ and DoD$^{Guided}$ performed best for all the datasets. Given the numerous good inter-epoch feature correspondences, we are able to effectively mitigate the \textit{dome} effect. DoD$^{Patch}$ performed slightly better than DoD$^{Guided}$ on  Fr{\'e}jus dataset, as the \textit{Patch} is more robust when extreme scene changes are \MajorMod{present}.
\paragraph{Ground displacements}
For ground displacements in Kobe, an up-lateral strike-slip movement along the sea is present in \textit{Guided} and \textit{Patch} (cf. Figure~\ref{DoDK}(i-j)), but not the other 3 methods. The observed signal is coherent with the fault of the Kobe earthquake known from independent data sources (cf. Figure~\ref{DoDK}(k)), according to which a striking north $30^{\circ}$-$60^{\circ}$ east rupture occurred along the northeast-southwest coastline across $ \sim $18 km~\cite{ian1996morphological}.\\
\paragraph{Check point accuracy}
For the check point accuracy in Pezenas and Fr{\'e}jus, the \textit{Patch} performed best, while the \textit{Guided} ranked second best for both datasets.
Also, the \textit{Patch} performed slightly better than \textit{Guided} for Pezenas, while the gap widened for Fr{\'e}jus, which is consistent with the result of DoDs.
\paragraph{Robustness}
Table~\ref{TiePtNumber} demonstrates the number of inter-epoch feature correspondences obtained with patch matching and guided matching before and after the 3D-RANSAC and cross correlation filters. Observe that 3D-RANSAC filter and cross correlation removed considerable number of feature correspondences, at the same time enough feature correspondences survived, which guaranteed robustness of our method. Moreover, the \textit{Patch} recovered considerably more feature correspondences than the \textit{Guided}, which is understandable as SuperGlue is more invariant over time than SIFT.\\
\begin{table}[htbp]
	\centering
	\begin{tabular}{||l|c|c|c|c|c|c|c||}\hline 
		&\multicolumn{3}{c|}{\textit{Patch}}&\multicolumn{3}{c|}{\textit{Guided}}\\\hline
		&tentative&enhanced&final&tentative&enhanced&final\\\hline\hline
		Pezenas& 1,754,434 & 1,686,486 & 284,067 & 1,671,818 & 390,438 & 129,674\\
		Fr{\'e}jus& 1,429,510 & 1,006,768 & 93,829 & 1,240,264 & 26,060 & 6,450\\
		Kobe& 616,630 & 147,200 & 12,671 & 54,098 & 4,484 & 1,785\\\hline
	\end{tabular}
	\caption{Number of inter-epoch feature correspondences in each step. Tentative stands for the feature correspondences resulting from the patch matching or the guided matching before applying the robust filters, enhanced stands for points after filtering with 3D-RANSAC, final means points after filtering with cross correlation.}
	\label{TiePtNumber}
\end{table}\\

\section{Conclusion}
\MajorMod{This work proposed to adopt DSMs and tiling schemes to get robust feature correspondences between historical images taken at different epochs. It opens up the possibility to unlock the potential of millions of historical images, which until now have been relatively unexplored due to lack of reliable processing techniques.}
Our matching strategy is able to extract reliable inter-epoch feature correspondences under significant scene changes where SIFT~\cite{lowe2004distinctive} and SuperGlue~\cite{sarlin2020superglue} fail.
Three sets of datasets are tested, including one that witnessed an earthquake. We showed that with the recovered feature correspondences we are able to mitigate the camera calibration systematic errors, leading to more accurate results, whether it is the geo-referenced DSM or ground displacements.

Our future work will {leverage} depth maps to train a neural network architecture in extracting robust features over time.

\section*{Acknowledgment}
This work was supported by ANR project DISRUPT (ANR-18-CE31-0012-0). We thank KIGAM (Korea Institute of Geoscience and Mineral Resources) for providing Kobe images.

	
	
	  \bibliographystyle{elsarticle-num-names} 
	  \bibliography{mybibfile}

\begin{thebibliography}{65}
\expandafter\ifx\csname natexlab\endcsname\relax\def\natexlab#1{#1}\fi
\providecommand{\url}[1]{\texttt{#1}}
\providecommand{\href}[2]{#2}
\providecommand{\path}[1]{#1}
\providecommand{\DOIprefix}{doi:}
\providecommand{\ArXivprefix}{arXiv:}
\providecommand{\URLprefix}{URL: }
\providecommand{\Pubmedprefix}{pmid:}
\providecommand{\doi}[1]{\href{http://dx.doi.org/#1}{\path{#1}}}
\providecommand{\Pubmed}[1]{\href{pmid:#1}{\path{#1}}}
\providecommand{\bibinfo}[2]{#2}
\ifx\xfnm\relax \def\xfnm[#1]{\unskip,\space#1}\fi
%Type = Article
\bibitem[{Cowley and Stichelbaut(2012)}]{cowley2012historic}
\bibinfo{author}{D.~C. Cowley}, \bibinfo{author}{B.~B. Stichelbaut},
\newblock \bibinfo{title}{Historic aerial photographic archives for european
  archaeology},
\newblock \bibinfo{journal}{European Journal of Archaeology}
  \bibinfo{volume}{15} (\bibinfo{year}{2012}) \bibinfo{pages}{217--236}.
%Type = Inproceedings
\bibitem[{Giordano and Mallet(2019)}]{sebastien2019archiving}
\bibinfo{author}{S.~Giordano}, \bibinfo{author}{C.~Mallet},
\newblock \bibinfo{title}{Archiving and geoprocessing of historical aerial
  images: current status in europe, official publication no 70},
\newblock in: \bibinfo{booktitle}{European Spatial Data Research},
  \bibinfo{year}{2019}.
%Type = Misc
\bibitem[{USGS(2019)}]{earthexplorer}
\bibinfo{author}{USGS}, \bibinfo{title}{{earthexplorer}},
  \bibinfo{howpublished}{https://earthexplorer.usgs.gov/},
  \bibinfo{year}{2019}.
%Type = Misc
\bibitem[{IGN(2019)}]{remonterletemps}
\bibinfo{author}{IGN}, \bibinfo{title}{{remonterletemps}},
  \bibinfo{howpublished}{https://remonterletemps.ign.fr/},
  \bibinfo{year}{2019}.
%Type = Article
\bibitem[{Lowe(2004)}]{lowe2004distinctive}
\bibinfo{author}{D.~G. Lowe},
\newblock \bibinfo{title}{Distinctive image features from scale-invariant
  keypoints},
\newblock \bibinfo{journal}{International journal of computer vision}
  \bibinfo{volume}{60} (\bibinfo{year}{2004}) \bibinfo{pages}{91--110}.
%Type = Misc
\bibitem[{MicMac(2021)}]{micmacGithub}
\bibinfo{author}{MicMac}, \bibinfo{title}{Github},
  \bibinfo{howpublished}{\url{https://github.com/micmacIGN/micmac/tree/master/src/uti_phgrm/TiePHistorical}},
  \bibinfo{year}{2021}.
%Type = Inproceedings
\bibitem[{Arandjelovi{\'c} and Zisserman(2012)}]{arandjelovic2012three}
\bibinfo{author}{R.~Arandjelovi{\'c}}, \bibinfo{author}{A.~Zisserman},
\newblock \bibinfo{title}{Three things everyone should know to improve object
  retrieval},
\newblock in: \bibinfo{booktitle}{2012 IEEE Conference on Computer Vision and
  Pattern Recognition}, \bibinfo{organization}{IEEE}, \bibinfo{year}{2012}, pp.
  \bibinfo{pages}{2911--2918}.
%Type = Inproceedings
\bibitem[{Bursuc et~al.(2015)Bursuc, Tolias, and J{\'e}gou}]{bursuc2015kernel}
\bibinfo{author}{A.~Bursuc}, \bibinfo{author}{G.~Tolias},
  \bibinfo{author}{H.~J{\'e}gou},
\newblock \bibinfo{title}{Kernel local descriptors with implicit rotation
  matching},
\newblock in: \bibinfo{booktitle}{Proceedings of the 5th ACM on International
  Conference on Multimedia Retrieval}, \bibinfo{year}{2015}, pp.
  \bibinfo{pages}{595--598}.
%Type = Inproceedings
\bibitem[{Dong and Soatto(2015)}]{dong2015domain}
\bibinfo{author}{J.~Dong}, \bibinfo{author}{S.~Soatto},
\newblock \bibinfo{title}{Domain-size pooling in local descriptors: Dsp-sift},
\newblock in: \bibinfo{booktitle}{Proceedings of the IEEE conference on
  computer vision and pattern recognition}, \bibinfo{year}{2015}, pp.
  \bibinfo{pages}{5097--5106}.
%Type = Article
\bibitem[{Yu and Morel(2011)}]{yu2011asift}
\bibinfo{author}{G.~Yu}, \bibinfo{author}{J.-M. Morel},
\newblock \bibinfo{title}{Asift: An algorithm for fully affine invariant
  comparison},
\newblock \bibinfo{journal}{Image Processing On Line} \bibinfo{volume}{1}
  (\bibinfo{year}{2011}) \bibinfo{pages}{11--38}.
%Type = Inproceedings
\bibitem[{Bay et~al.(2006)Bay, Tuytelaars, and Van~Gool}]{bay2006surf}
\bibinfo{author}{H.~Bay}, \bibinfo{author}{T.~Tuytelaars},
  \bibinfo{author}{L.~Van~Gool},
\newblock \bibinfo{title}{Surf: Speeded up robust features},
\newblock in: \bibinfo{booktitle}{European conference on computer vision},
  \bibinfo{year}{2006}, pp. \bibinfo{pages}{404--417}.
%Type = Article
\bibitem[{Mikolajczyk and Schmid(2004)}]{mikolajczyk2004scale}
\bibinfo{author}{K.~Mikolajczyk}, \bibinfo{author}{C.~Schmid},
\newblock \bibinfo{title}{Scale \& affine invariant interest point detectors},
\newblock \bibinfo{journal}{International journal of computer vision}
  \bibinfo{volume}{60} (\bibinfo{year}{2004}) \bibinfo{pages}{63--86}.
%Type = Inproceedings
\bibitem[{Alcantarilla et~al.(2012)Alcantarilla, Bartoli, and
  Davison}]{alcantarilla2012kaze}
\bibinfo{author}{P.~F. Alcantarilla}, \bibinfo{author}{A.~Bartoli},
  \bibinfo{author}{A.~J. Davison},
\newblock \bibinfo{title}{Kaze features},
\newblock in: \bibinfo{booktitle}{European Conference on Computer Vision},
  \bibinfo{year}{2012}, pp. \bibinfo{pages}{214--227}.
%Type = Inproceedings
\bibitem[{Yi et~al.(2016)Yi, Trulls, Lepetit, and Fua}]{yi2016lift}
\bibinfo{author}{K.~M. Yi}, \bibinfo{author}{E.~Trulls},
  \bibinfo{author}{V.~Lepetit}, \bibinfo{author}{P.~Fua},
\newblock \bibinfo{title}{Lift: Learned invariant feature transform},
\newblock in: \bibinfo{booktitle}{European Conference on Computer Vision},
  \bibinfo{year}{2016}, pp. \bibinfo{pages}{467--483}.
%Type = Inproceedings
\bibitem[{Tian et~al.(2017)Tian, Fan, and Wu}]{tian2017l2}
\bibinfo{author}{Y.~Tian}, \bibinfo{author}{B.~Fan}, \bibinfo{author}{F.~Wu},
\newblock \bibinfo{title}{L2-net: Deep learning of discriminative patch
  descriptor in euclidean space},
\newblock in: \bibinfo{booktitle}{Proceedings of the IEEE Conference on
  Computer Vision and Pattern Recognition}, \bibinfo{year}{2017}, pp.
  \bibinfo{pages}{661--669}.
%Type = Inproceedings
\bibitem[{Mishchuk et~al.(2017)Mishchuk, Mishkin, Radenovic, and
  Matas}]{mishchuk2017working}
\bibinfo{author}{A.~Mishchuk}, \bibinfo{author}{D.~Mishkin},
  \bibinfo{author}{F.~Radenovic}, \bibinfo{author}{J.~Matas},
\newblock \bibinfo{title}{Working hard to know your neighbor's margins: Local
  descriptor learning loss},
\newblock in: \bibinfo{booktitle}{Advances in Neural Information Processing
  Systems}, \bibinfo{year}{2017}, pp. \bibinfo{pages}{4826--4837}.
%Type = Inproceedings
\bibitem[{Luo et~al.(2019)Luo, Shen, Zhou, Zhang, Yao, Li, Fang, and
  Quan}]{luo2019contextdesc}
\bibinfo{author}{Z.~Luo}, \bibinfo{author}{T.~Shen}, \bibinfo{author}{L.~Zhou},
  \bibinfo{author}{J.~Zhang}, \bibinfo{author}{Y.~Yao},
  \bibinfo{author}{S.~Li}, \bibinfo{author}{T.~Fang},
  \bibinfo{author}{L.~Quan},
\newblock \bibinfo{title}{Contextdesc: Local descriptor augmentation with
  cross-modality context},
\newblock in: \bibinfo{booktitle}{Proceedings of the IEEE Conference on
  Computer Vision and Pattern Recognition}, \bibinfo{year}{2019}, pp.
  \bibinfo{pages}{2527--2536}.
%Type = Inproceedings
\bibitem[{Ono et~al.(2018)Ono, Trulls, Fua, and Yi}]{ono2018lf}
\bibinfo{author}{Y.~Ono}, \bibinfo{author}{E.~Trulls},
  \bibinfo{author}{P.~Fua}, \bibinfo{author}{K.~M. Yi},
\newblock \bibinfo{title}{Lf-net: learning local features from images},
\newblock in: \bibinfo{booktitle}{Advances in Neural Information Processing
  Systems}, \bibinfo{year}{2018}, pp. \bibinfo{pages}{6234--6244}.
%Type = Inproceedings
\bibitem[{Noh et~al.(2017)Noh, Araujo, Sim, Weyand, and Han}]{noh2017DELF}
\bibinfo{author}{H.~Noh}, \bibinfo{author}{A.~Araujo},
  \bibinfo{author}{J.~Sim}, \bibinfo{author}{T.~Weyand},
  \bibinfo{author}{B.~Han},
\newblock \bibinfo{title}{Large-scale image retrieval with attentive deep local
  features},
\newblock in: \bibinfo{booktitle}{Proceedings of the IEEE international
  conference on computer vision}, \bibinfo{year}{2017}, pp.
  \bibinfo{pages}{3456--3465}.
%Type = Inproceedings
\bibitem[{DeTone et~al.(2018)DeTone, Malisiewicz, and
  Rabinovich}]{detone2018superpoint}
\bibinfo{author}{D.~DeTone}, \bibinfo{author}{T.~Malisiewicz},
  \bibinfo{author}{A.~Rabinovich},
\newblock \bibinfo{title}{Superpoint: Self-supervised interest point detection
  and description},
\newblock in: \bibinfo{booktitle}{Proceedings of the IEEE Conference on
  Computer Vision and Pattern Recognition Workshops}, \bibinfo{year}{2018}, pp.
  \bibinfo{pages}{224--236}.
%Type = Inproceedings
\bibitem[{Dusmanu et~al.(2019)Dusmanu, Rocco, Pajdla, Pollefeys, Sivic, Torii,
  and Sattler}]{dusmanu2019d2}
\bibinfo{author}{M.~Dusmanu}, \bibinfo{author}{I.~Rocco},
  \bibinfo{author}{T.~Pajdla}, \bibinfo{author}{M.~Pollefeys},
  \bibinfo{author}{J.~Sivic}, \bibinfo{author}{A.~Torii},
  \bibinfo{author}{T.~Sattler},
\newblock \bibinfo{title}{D2-net: A trainable cnn for joint detection and
  description of local features},
\newblock in: \bibinfo{booktitle}{2019 IEEE Conference on Computer Vision and
  Pattern Recognition}, \bibinfo{year}{2019}, pp. \bibinfo{pages}{8092--8101}.
%Type = Inproceedings
\bibitem[{Luo et~al.(2020)Luo, Zhou, Bai, Chen, Zhang, Yao, Li, Fang, and
  Quan}]{luo2020aslfeat}
\bibinfo{author}{Z.~Luo}, \bibinfo{author}{L.~Zhou}, \bibinfo{author}{X.~Bai},
  \bibinfo{author}{H.~Chen}, \bibinfo{author}{J.~Zhang},
  \bibinfo{author}{Y.~Yao}, \bibinfo{author}{S.~Li}, \bibinfo{author}{T.~Fang},
  \bibinfo{author}{L.~Quan},
\newblock \bibinfo{title}{Aslfeat: Learning local features of accurate shape
  and localization},
\newblock in: \bibinfo{booktitle}{Proceedings of the IEEE/CVF Conference on
  Computer Vision and Pattern Recognition}, \bibinfo{year}{2020}, pp.
  \bibinfo{pages}{6589--6598}.
%Type = Inproceedings
\bibitem[{Revaud et~al.(2019)Revaud, De~Souza, Humenberger, and
  Weinzaepfel}]{revaud2019r2d2}
\bibinfo{author}{J.~Revaud}, \bibinfo{author}{C.~De~Souza},
  \bibinfo{author}{M.~Humenberger}, \bibinfo{author}{P.~Weinzaepfel},
\newblock \bibinfo{title}{R2d2: Reliable and repeatable detector and
  descriptor},
\newblock in: \bibinfo{booktitle}{Advances in Neural Information Processing
  Systems}, \bibinfo{year}{2019}, pp. \bibinfo{pages}{12405--12415}.
%Type = Article
\bibitem[{Wiles et~al.(2020)Wiles, Ehrhardt, and Zisserman}]{wiles2020d2d}
\bibinfo{author}{O.~Wiles}, \bibinfo{author}{S.~Ehrhardt},
  \bibinfo{author}{A.~Zisserman},
\newblock \bibinfo{title}{D2d: Learning to find good correspondences for image
  matching and manipulation},
\newblock \bibinfo{journal}{arXiv preprint arXiv:2007.08480}
  (\bibinfo{year}{2020}).
%Type = Inproceedings
\bibitem[{Sarlin et~al.(2020)Sarlin, DeTone, Malisiewicz, and
  Rabinovich}]{sarlin2020superglue}
\bibinfo{author}{P.-E. Sarlin}, \bibinfo{author}{D.~DeTone},
  \bibinfo{author}{T.~Malisiewicz}, \bibinfo{author}{A.~Rabinovich},
\newblock \bibinfo{title}{Superglue: Learning feature matching with graph
  neural networks},
\newblock in: \bibinfo{booktitle}{Proceedings of the IEEE/CVF Conference on
  Computer Vision and Pattern Recognition}, \bibinfo{year}{2020}, pp.
  \bibinfo{pages}{4938--4947}.
%Type = Inproceedings
\bibitem[{Schonberger et~al.(2017)Schonberger, Hardmeier, Sattler, and
  Pollefeys}]{schonberger2017comparative}
\bibinfo{author}{J.~L. Schonberger}, \bibinfo{author}{H.~Hardmeier},
  \bibinfo{author}{T.~Sattler}, \bibinfo{author}{M.~Pollefeys},
\newblock \bibinfo{title}{Comparative evaluation of hand-crafted and learned
  local features},
\newblock in: \bibinfo{booktitle}{Proceedings of the IEEE Conference on
  Computer Vision and Pattern Recognition}, \bibinfo{year}{2017}, pp.
  \bibinfo{pages}{1482--1491}.
%Type = Article
\bibitem[{Jin et~al.(2020)Jin, Mishkin, Mishchuk, Matas, Fua, Yi, and
  Trulls}]{jin2020image}
\bibinfo{author}{Y.~Jin}, \bibinfo{author}{D.~Mishkin},
  \bibinfo{author}{A.~Mishchuk}, \bibinfo{author}{J.~Matas},
  \bibinfo{author}{P.~Fua}, \bibinfo{author}{K.~M. Yi},
  \bibinfo{author}{E.~Trulls},
\newblock \bibinfo{title}{Image matching across wide baselines: From paper to
  practice},
\newblock \bibinfo{journal}{2020 IEEE Conference on Computer Vision and Pattern
  Recognition}  (\bibinfo{year}{2020}).
%Type = Misc
\bibitem[{Trulls et~al.(2020)Trulls, Jin, Yi, Mishkin, Matas, and
  Fua}]{imagematchingchallenge2020}
\bibinfo{author}{E.~Trulls}, \bibinfo{author}{Y.~Jin}, \bibinfo{author}{K.~M.
  Yi}, \bibinfo{author}{D.~Mishkin}, \bibinfo{author}{J.~Matas},
  \bibinfo{author}{P.~Fua}, \bibinfo{title}{{Image Matching Challenge 2020}},
  \bibinfo{howpublished}{https://vision.uvic.ca/image-matching-challenge/},
  \bibinfo{year}{2020}.
%Type = Article
\bibitem[{Pinto et~al.(2019)Pinto, Gon{\c{c}}alves, Beja, and
  Pradinho~Honrado}]{pinto2019archived}
\bibinfo{author}{A.~T. Pinto}, \bibinfo{author}{J.~A. Gon{\c{c}}alves},
  \bibinfo{author}{P.~Beja}, \bibinfo{author}{J.~Pradinho~Honrado},
\newblock \bibinfo{title}{From archived historical aerial imagery to
  informative orthophotos: A framework for retrieving the past in long-term
  socioecological research},
\newblock \bibinfo{journal}{Remote Sensing} \bibinfo{volume}{11}
  (\bibinfo{year}{2019}) \bibinfo{pages}{1388}.
%Type = Article
\bibitem[{Bo{\.z}ek et~al.(2019)Bo{\.z}ek, Janus, and
  Mitka}]{bozek2019analysis}
\bibinfo{author}{P.~Bo{\.z}ek}, \bibinfo{author}{J.~Janus},
  \bibinfo{author}{B.~Mitka},
\newblock \bibinfo{title}{Analysis of changes in forest structure using point
  clouds from historical aerial photographs},
\newblock \bibinfo{journal}{Remote Sensing} \bibinfo{volume}{11}
  (\bibinfo{year}{2019}) \bibinfo{pages}{2259}.
%Type = Article
\bibitem[{Persia et~al.(2020)Persia, Barca, Greco, Marzulli, and
  Tartarino}]{persia2020archival}
\bibinfo{author}{M.~Persia}, \bibinfo{author}{E.~Barca},
  \bibinfo{author}{R.~Greco}, \bibinfo{author}{M.~Marzulli},
  \bibinfo{author}{P.~Tartarino},
\newblock \bibinfo{title}{Archival aerial images georeferencing: A
  geostatistically-based approach for improving orthophoto accuracy with
  minimal number of ground control points},
\newblock \bibinfo{journal}{Remote Sensing} \bibinfo{volume}{12}
  (\bibinfo{year}{2020}) \bibinfo{pages}{2232}.
%Type = Article
\bibitem[{Micheletti et~al.(2015)Micheletti, Lane, and
  Chandler}]{micheletti2015application}
\bibinfo{author}{N.~Micheletti}, \bibinfo{author}{S.~N. Lane},
  \bibinfo{author}{J.~H. Chandler},
\newblock \bibinfo{title}{Application of archival aerial photogrammetry to
  quantify climate forcing of alpine landscapes},
\newblock \bibinfo{journal}{The Photogrammetric Record} \bibinfo{volume}{30}
  (\bibinfo{year}{2015}) \bibinfo{pages}{143--165}.
%Type = Article
\bibitem[{M{\"o}lg and Bolch(2017)}]{molg2017structure}
\bibinfo{author}{N.~M{\"o}lg}, \bibinfo{author}{T.~Bolch},
\newblock \bibinfo{title}{Structure-from-motion using historical aerial images
  to analyse changes in glacier surface elevation},
\newblock \bibinfo{journal}{Remote Sensing} \bibinfo{volume}{9}
  (\bibinfo{year}{2017}) \bibinfo{pages}{1021}.
%Type = Article
\bibitem[{Giordano et~al.(2018)Giordano, Le~Bris, and
  Mallet}]{giordano2018toward}
\bibinfo{author}{S.~Giordano}, \bibinfo{author}{A.~Le~Bris},
  \bibinfo{author}{C.~Mallet},
\newblock \bibinfo{title}{Toward automatic georeferencing of archival aerial
  photogrammetric surveys},
\newblock \bibinfo{journal}{ISPRS Annals of Photogrammetry, Remote Sensing and
  Spatial Information Sciences} \bibinfo{volume}{IV-2} (\bibinfo{year}{2018})
  \bibinfo{pages}{105--112}.
%Type = Inproceedings
\bibitem[{Dalal and Triggs(2005)}]{dalal2005histograms}
\bibinfo{author}{N.~Dalal}, \bibinfo{author}{B.~Triggs},
\newblock \bibinfo{title}{Histograms of oriented gradients for human
  detection},
\newblock in: \bibinfo{booktitle}{2005 IEEE computer society conference on
  computer vision and pattern recognition (CVPR'05)},
  volume~\bibinfo{volume}{1}, \bibinfo{organization}{Ieee},
  \bibinfo{year}{2005}, pp. \bibinfo{pages}{886--893}.
%Type = Article
\bibitem[{Feurer and Vinatier(2018)}]{feurer2018joining}
\bibinfo{author}{D.~Feurer}, \bibinfo{author}{F.~Vinatier},
\newblock \bibinfo{title}{Joining multi-epoch archival aerial images in a
  single sfm block allows 3-d change detection with almost exclusively image
  information},
\newblock \bibinfo{journal}{ISPRS journal of photogrammetry and remote sensing}
  \bibinfo{volume}{146} (\bibinfo{year}{2018}) \bibinfo{pages}{495--506}.
%Type = Article
\bibitem[{Filhol et~al.(2019)Filhol, Perret, Girod, Sutter, Schuler, and
  Burkhart}]{filhol2019time}
\bibinfo{author}{S.~Filhol}, \bibinfo{author}{A.~Perret},
  \bibinfo{author}{L.~Girod}, \bibinfo{author}{G.~Sutter},
  \bibinfo{author}{T.~Schuler}, \bibinfo{author}{J.~Burkhart},
\newblock \bibinfo{title}{Time-lapse photogrammetry of distributed snow depth
  during snowmelt},
\newblock \bibinfo{journal}{Water Resources Research} \bibinfo{volume}{55}
  (\bibinfo{year}{2019}) \bibinfo{pages}{7916--7926}.
%Type = Article
\bibitem[{Cook and Dietze(2019)}]{cook2019simple}
\bibinfo{author}{K.~L. Cook}, \bibinfo{author}{M.~Dietze},
\newblock \bibinfo{title}{A simple workflow for robust low-cost uav-derived
  change detection without ground control points},
\newblock \bibinfo{journal}{Earth Surface Dynamics} \bibinfo{volume}{7}
  (\bibinfo{year}{2019}) \bibinfo{pages}{1009--1017}.
%Type = Article
\bibitem[{Parente et~al.(2021)Parente, Chandler, and
  Dixon}]{parente2021automated}
\bibinfo{author}{L.~Parente}, \bibinfo{author}{J.~H. Chandler},
  \bibinfo{author}{N.~Dixon},
\newblock \bibinfo{title}{Automated registration of sfm-mvs multitemporal
  datasets using terrestrial and oblique aerial images},
\newblock \bibinfo{journal}{The Photogrammetric Record} \bibinfo{volume}{36}
  (\bibinfo{year}{2021}) \bibinfo{pages}{12--35}.
%Type = Article
\bibitem[{Blanch et~al.(2021)Blanch, Eltner, Guinau, and
  Abellan}]{blanch2021multi}
\bibinfo{author}{X.~Blanch}, \bibinfo{author}{A.~Eltner},
  \bibinfo{author}{M.~Guinau}, \bibinfo{author}{A.~Abellan},
\newblock \bibinfo{title}{Multi-epoch and multi-imagery (memi) photogrammetric
  workflow for enhanced change detection using time-lapse cameras},
\newblock \bibinfo{journal}{Remote Sensing} \bibinfo{volume}{13}
  (\bibinfo{year}{2021}) \bibinfo{pages}{1460}.
%Type = Inproceedings
\bibitem[{Zhang et~al.(2020)Zhang, Rupnik, and
  Pierrot-Deseilligny}]{zhang2020guided}
\bibinfo{author}{L.~Zhang}, \bibinfo{author}{E.~Rupnik},
  \bibinfo{author}{M.~Pierrot-Deseilligny},
\newblock \bibinfo{title}{Guided feature matching for multi-epoch historical
  image blocks pose estimation},
\newblock in: \bibinfo{booktitle}{ISPRS Ann. Photogramm. Remote Sens. Spatial
  Inf. Sci.}, \bibinfo{year}{2020}.
%Type = Article
\bibitem[{Maiwald and Maas(2021)}]{maiwald2021automatic}
\bibinfo{author}{F.~Maiwald}, \bibinfo{author}{H.-G. Maas},
\newblock \bibinfo{title}{An automatic workflow for orientation of historical
  images with large radiometric and geometric differences},
\newblock \bibinfo{journal}{The Photogrammetric Record}
  (\bibinfo{year}{2021}).
%Type = Article
\bibitem[{Beltrami et~al.(2019)Beltrami, Cavezzali, Chiabrando,
  Iaccarino~Idelson, Patrucco, and Rinaudo}]{beltrami20193d}
\bibinfo{author}{C.~Beltrami}, \bibinfo{author}{D.~Cavezzali},
  \bibinfo{author}{F.~Chiabrando}, \bibinfo{author}{A.~Iaccarino~Idelson},
  \bibinfo{author}{G.~Patrucco}, \bibinfo{author}{F.~Rinaudo},
\newblock \bibinfo{title}{3d digital and physical reconstruction of a collapsed
  dome using sfm techniques from historical images.},
\newblock \bibinfo{journal}{International Archives of the Photogrammetry,
  Remote Sensing \& Spatial Information Sciences}  (\bibinfo{year}{2019}).
%Type = Article
\bibitem[{Bevilacqua et~al.(2019)Bevilacqua, Caroti, Piemonte, and
  Ulivieri}]{bevilacqua2019reconstruction}
\bibinfo{author}{M.~Bevilacqua}, \bibinfo{author}{G.~Caroti},
  \bibinfo{author}{A.~Piemonte}, \bibinfo{author}{D.~Ulivieri},
\newblock \bibinfo{title}{Reconstruction of lost architectural volumes by
  integration of photogrammetry from archive imagery with 3d models of the
  status quo.},
\newblock \bibinfo{journal}{International Archives of the Photogrammetry,
  Remote Sensing \& Spatial Information Sciences}  (\bibinfo{year}{2019}).
%Type = Article
\bibitem[{Maiwald(2019)}]{maiwald2019generation}
\bibinfo{author}{F.~Maiwald},
\newblock \bibinfo{title}{Generation of a benchmark dataset using historical
  photographs for an automated evaluation of different feature matching
  methods.},
\newblock \bibinfo{journal}{International Archives of the Photogrammetry,
  Remote Sensing \& Spatial Information Sciences}  (\bibinfo{year}{2019}).
%Type = Article
\bibitem[{Fischler and Bolles(1981)}]{fischler1981random}
\bibinfo{author}{M.~A. Fischler}, \bibinfo{author}{R.~C. Bolles},
\newblock \bibinfo{title}{Random sample consensus: a paradigm for model fitting
  with applications to image analysis and automated cartography},
\newblock \bibinfo{journal}{Communications of the ACM} \bibinfo{volume}{24}
  (\bibinfo{year}{1981}) \bibinfo{pages}{381--395}.
%Type = Book
\bibitem[{Sonka et~al.(2014)Sonka, Hlavac, and Boyle}]{sonka2014image}
\bibinfo{author}{M.~Sonka}, \bibinfo{author}{V.~Hlavac},
  \bibinfo{author}{R.~Boyle}, \bibinfo{title}{Image processing, analysis, and
  machine vision}, \bibinfo{publisher}{Cengage Learning}, \bibinfo{year}{2014}.
%Type = Article
\bibitem[{Leroy and Rousseeuw(1987)}]{leroy1987robust}
\bibinfo{author}{A.~M. Leroy}, \bibinfo{author}{P.~J. Rousseeuw},
\newblock \bibinfo{title}{Robust regression and outlier detection},
\newblock \bibinfo{journal}{Wiley}  (\bibinfo{year}{1987}).
%Type = Article
\bibitem[{Torr and Zisserman(2000)}]{torr2000mlesac}
\bibinfo{author}{P.~H. Torr}, \bibinfo{author}{A.~Zisserman},
\newblock \bibinfo{title}{Mlesac: A new robust estimator with application to
  estimating image geometry},
\newblock \bibinfo{journal}{Computer vision and image understanding}
  \bibinfo{volume}{78} (\bibinfo{year}{2000}) \bibinfo{pages}{138--156}.
%Type = Inproceedings
\bibitem[{Chum and Matas(2005)}]{chum2005matching}
\bibinfo{author}{O.~Chum}, \bibinfo{author}{J.~Matas},
\newblock \bibinfo{title}{Matching with prosac-progressive sample consensus},
\newblock in: \bibinfo{booktitle}{2005 IEEE computer society conference on
  computer vision and pattern recognition (CVPR'05)},
  volume~\bibinfo{volume}{1}, \bibinfo{organization}{IEEE},
  \bibinfo{year}{2005}, pp. \bibinfo{pages}{220--226}.
%Type = Inproceedings
\bibitem[{Chum et~al.(2005)Chum, Werner, and Matas}]{chum2005two}
\bibinfo{author}{O.~Chum}, \bibinfo{author}{T.~Werner},
  \bibinfo{author}{J.~Matas},
\newblock \bibinfo{title}{Two-view geometry estimation unaffected by a dominant
  plane},
\newblock in: \bibinfo{booktitle}{2005 IEEE Computer Society Conference on
  Computer Vision and Pattern Recognition (CVPR'05)},
  volume~\bibinfo{volume}{1}, \bibinfo{organization}{IEEE},
  \bibinfo{year}{2005}, pp. \bibinfo{pages}{772--779}.
%Type = Inproceedings
\bibitem[{Barath and Matas(2018)}]{barath2018graph}
\bibinfo{author}{D.~Barath}, \bibinfo{author}{J.~Matas},
\newblock \bibinfo{title}{Graph-cut ransac},
\newblock in: \bibinfo{booktitle}{Proceedings of the IEEE Conference on
  Computer Vision and Pattern Recognition}, \bibinfo{year}{2018}, pp.
  \bibinfo{pages}{6733--6741}.
%Type = Inproceedings
\bibitem[{Barath et~al.(2019)Barath, Matas, and Noskova}]{barath2019magsac}
\bibinfo{author}{D.~Barath}, \bibinfo{author}{J.~Matas},
  \bibinfo{author}{J.~Noskova},
\newblock \bibinfo{title}{Magsac: marginalizing sample consensus},
\newblock in: \bibinfo{booktitle}{Proceedings of the IEEE Conference on
  Computer Vision and Pattern Recognition}, \bibinfo{year}{2019}, pp.
  \bibinfo{pages}{10197--10205}.
%Type = Inproceedings
\bibitem[{Brachmann et~al.(2017)Brachmann, Krull, Nowozin, Shotton, Michel,
  Gumhold, and Rother}]{brachmann2017dsac}
\bibinfo{author}{E.~Brachmann}, \bibinfo{author}{A.~Krull},
  \bibinfo{author}{S.~Nowozin}, \bibinfo{author}{J.~Shotton},
  \bibinfo{author}{F.~Michel}, \bibinfo{author}{S.~Gumhold},
  \bibinfo{author}{C.~Rother},
\newblock \bibinfo{title}{Dsac-differentiable ransac for camera localization},
\newblock in: \bibinfo{booktitle}{Proceedings of the IEEE Conference on
  Computer Vision and Pattern Recognition}, \bibinfo{year}{2017}, pp.
  \bibinfo{pages}{6684--6692}.
%Type = Inproceedings
\bibitem[{Moo~Yi et~al.(2018)Moo~Yi, Trulls, Ono, Lepetit, Salzmann, and
  Fua}]{moo2018learning}
\bibinfo{author}{K.~Moo~Yi}, \bibinfo{author}{E.~Trulls},
  \bibinfo{author}{Y.~Ono}, \bibinfo{author}{V.~Lepetit},
  \bibinfo{author}{M.~Salzmann}, \bibinfo{author}{P.~Fua},
\newblock \bibinfo{title}{Learning to find good correspondences},
\newblock in: \bibinfo{booktitle}{Proceedings of the IEEE Conference on
  Computer Vision and Pattern Recognition}, \bibinfo{year}{2018}, pp.
  \bibinfo{pages}{2666--2674}.
%Type = Article
\bibitem[{Pierrot-Deseilligny and Cl{\'e}ry(2012)}]{deseilligny2011apero}
\bibinfo{author}{M.~Pierrot-Deseilligny}, \bibinfo{author}{I.~Cl{\'e}ry},
\newblock \bibinfo{title}{Apero, an open source bundle adjusment software for
  automatic calibration and orientation of set of images},
\newblock \bibinfo{journal}{ISPRS International Archives of the Photogrammetry,
  Remote Sensing and Spatial Information Sciences}
  \bibinfo{volume}{{XXXVIII}-5/W16} (\bibinfo{year}{2012})
  \bibinfo{pages}{269--276}.
%Type = Inproceedings
\bibitem[{Schonberger and Frahm(2016)}]{schonberger2016structure}
\bibinfo{author}{J.~L. Schonberger}, \bibinfo{author}{J.-M. Frahm},
\newblock \bibinfo{title}{Structure-from-motion revisited},
\newblock in: \bibinfo{booktitle}{Proceedings of the IEEE Conference on
  Computer Vision and Pattern Recognition}, \bibinfo{year}{2016}, pp.
  \bibinfo{pages}{4104--4113}.
%Type = Misc
\bibitem[{Moulon et~al.(2016)Moulon, Monasse, Marlet, and Others}]{openMVG}
\bibinfo{author}{P.~Moulon}, \bibinfo{author}{P.~Monasse},
  \bibinfo{author}{R.~Marlet}, \bibinfo{author}{Others},
  \bibinfo{title}{Openmvg},
  \bibinfo{howpublished}{\url{https://github.com/openMVG/openMVG}},
  \bibinfo{year}{2016}.
%Type = Misc
\bibitem[{Sweeney(2015)}]{theia}
\bibinfo{author}{C.~Sweeney}, \bibinfo{title}{Theia multiview geometry library:
  Tutorial \& reference}, \bibinfo{howpublished}{\url{http://theia-sfm.org}},
  \bibinfo{year}{2015}.
%Type = Article
\bibitem[{Pierrot-Deseilligny and Paparoditis(2006)}]{mpd:06:sgm}
\bibinfo{author}{M.~Pierrot-Deseilligny}, \bibinfo{author}{N.~Paparoditis},
\newblock \bibinfo{title}{A multiresolution and optimization-based image
  matching approach: An application to surface reconstruction from spot5-hrs
  stereo imagery},
\newblock \bibinfo{journal}{Archives of Photogrammetry, Remote Sensing and
  Spatial Information Sciences} \bibinfo{volume}{36} (\bibinfo{year}{2006})
  \bibinfo{pages}{1--5}.
%Type = Article
\bibitem[{Fraser(1997)}]{fraser1997digital}
\bibinfo{author}{C.~S. Fraser},
\newblock \bibinfo{title}{Digital camera self-calibration},
\newblock \bibinfo{journal}{ISPRS Journal of Photogrammetry and Remote sensing}
  \bibinfo{volume}{52} (\bibinfo{year}{1997}) \bibinfo{pages}{149--159}.
%Type = Article
\bibitem[{Pierrot-Deseilligny et~al.(2015)Pierrot-Deseilligny, Rupnik, Girod,
  Belvaux, Maillet, Deveau, and Choqueux}]{marc2016micmac}
\bibinfo{author}{M.~Pierrot-Deseilligny}, \bibinfo{author}{E.~Rupnik},
  \bibinfo{author}{L.~Girod}, \bibinfo{author}{J.~Belvaux},
  \bibinfo{author}{G.~Maillet}, \bibinfo{author}{M.~Deveau},
  \bibinfo{author}{G.~Choqueux},
\newblock \bibinfo{title}{Micmac, apero, pastis and other beverages in a
  nutshell},
\newblock \bibinfo{journal}{MicMac documentation} \bibinfo{volume}{4}
  (\bibinfo{year}{2015}).
%Type = Article
\bibitem[{Souchon et~al.(2010)Souchon, Thom, Meynard, Martin, and
  Pierrot-Deseilligny}]{souchon2010ign}
\bibinfo{author}{J.-P. Souchon}, \bibinfo{author}{C.~Thom},
  \bibinfo{author}{C.~Meynard}, \bibinfo{author}{O.~Martin},
  \bibinfo{author}{M.~Pierrot-Deseilligny},
\newblock \bibinfo{title}{The ign camv2 system},
\newblock \bibinfo{journal}{The Photogrammetric Record} \bibinfo{volume}{25}
  (\bibinfo{year}{2010}) \bibinfo{pages}{402--421}.
%Type = Article
\bibitem[{Rosu et~al.(2015)Rosu, Pierrot-Deseilligny, Delorme, Binet, and
  Klinger}]{rosu2015measurement}
\bibinfo{author}{A.~Rosu}, \bibinfo{author}{M.~Pierrot-Deseilligny},
  \bibinfo{author}{A.~Delorme}, \bibinfo{author}{R.~Binet},
  \bibinfo{author}{Y.~Klinger},
\newblock \bibinfo{title}{Measurement of ground displacement from optical
  satellite image correlation using the free open-source software micmac},
\newblock \bibinfo{journal}{ISPRS Journal of Photogrammetry and Remote Sensing}
  \bibinfo{volume}{100} (\bibinfo{year}{2015}) \bibinfo{pages}{48--59}.
%Type = Article
\bibitem[{Ian and Uda(1996)}]{ian1996morphological}
\bibinfo{author}{A.~Ian}, \bibinfo{author}{S.~Uda},
\newblock \bibinfo{title}{Morphological characteristics of the earthquake
  surface ruptures on awaji island, associated with the 1995 southern hyogo
  prefecture earthquake},
\newblock \bibinfo{journal}{Island Arc} \bibinfo{volume}{5}
  (\bibinfo{year}{1996}) \bibinfo{pages}{1--15}.

\end{thebibliography}
	
	
%	\begin{thebibliography}{00}
%		
%		%% \bibitem[Author(year)]{label}
%		%% Text of bibliographic item
%		
%		\bibitem[ ()]{}
%		
%	\end{thebibliography}

\end{document}